\documentclass[pdflatex]{sn-jnl}

\usepackage{amssymb}
\usepackage{amsmath}
\usepackage{graphicx}
\usepackage{braket}
\usepackage{bm}
\usepackage{tikz}

\usepackage{quantikz}
\usepackage[dvipsnames]{xcolor}
\usepackage{tabularx}
\usepackage{caption}
\usepackage{wrapfig}
\usepackage{rotating}
\usepackage{url}
\usepackage{geometry}
\usepackage{setspace}
\usepackage{adjustbox}
\usepackage[numbers, sort&compress]{natbib}

\newcolumntype{L}[1]{>{\hsize=#1\hsize\raggedright\arraybackslash}X}%
\graphicspath{ {./Figures/} }

\begin{document}

\title[Large-Scale Benchmarking of Quantum Neural Network Configurations for Financial Time Series Forecasting]{Large-Scale Benchmarking of Quantum Neural Network Configurations for Financial Time Series Forecasting}

\author[1]{\fnm{Jack} \sur{Waller}}\email{k2407357@kingston.ac.uk}

\author*[1]{\fnm{Xing} \sur{Liang}}\email{X.Liang@kingston.ac.uk}

\author[1]{\fnm{Dimitrios} \sur{Makris}}\email{D.Makris@kingston.ac.uk}

\author[2]{\fnm{Rajagopal} \sur{Nilavalan}}\email{Nila.Nilavalan@brunel.ac.uk}

\affil*[1]{\orgdiv{School of Computer Science and Mathematics}, \orgname{Kingston University London}, \orgaddress{\city{London}, \postcode{KT1 2EE}, \country{U.K.}}}

\affil[2]{\orgdiv{Department of Electronics and Electrical Engineering}, \orgname{Brunel University of London}, \orgaddress{\city{London}, \postcode{UB8 3PH}, \country{U.K.}}}

\abstract{Quantum machine learning, and quantum neural networks (QNNs) in particular, are advancing fields with growing potential. Although systematic comparisons of QNN configurations have been explored primarily for classification tasks, comparatively little attention has been given to regression problems, particularly financial time series forecasting. This study presents a large-scale systematic comparative evaluation of QNN component configurations for financial time series forecasting, using the GBP/USD spot exchange rate as a case study. A grid search across encoding methods, ansatz designs, qubit counts, layer depths, and cost functions yields 1,368 distinct model configurations, each evaluated in terms of prediction accuracy, computational cost, and convergence behaviour. The results reveal unique insights into how the choice of methods influences performance, such as that gate selection and arrangement are more critical to model success than raw parameter count, and that entanglement is a system-level property of the full circuit rather than solely at the ansatz level. The best-performing QNN configuration achieves an $R^2$ score of 0.985, outperforming a classical BiLSTM baseline. Additionally, the impact of real quantum hardware noise is assessed through execution on the IQM Emerald device, revealing that gate errors and decoherence represent a significant barrier to practical deployment, with gate selection and circuit depth identified as key determinants of hardware noise resilience. Overall, the findings provide practical architectural guidance for QNN design and establish a baseline characterisation of QNN noise sensitivity on near-term quantum devices.}


\maketitle
\clearpage
\section{Introduction}
Machine learning approaches have demonstrated significant advances across numerous domains, achieving breakthroughs including medical diagnosis \cite{407Rajpurkar}\cite{408Sadr}, autonomous systems \cite{409Kiran}\cite{410Chen}, and computer vision \cite{424Wang}. Within this broad landscape, neural networks have proven particularly adept at temporal pattern recognition, showing strong performance in time series forecasting where sequential dependencies and complex temporal relationships must be captured \cite{349Tang}\cite{425Kim}. However, these same complex temporal relationships expose limitations in classical models, which can struggle to efficiently represent the intricate correlations and interdependencies present in high-dimensional data \cite{411Vhaduri}\cite{412Sodar}.

Quantum computing has emerged as a rapidly advancing field, with recent developments in the noisy intermediate-scale quantum devices (NISQ) era \cite{138Preskill} opening new opportunities for practical quantum applications. These advances have demonstrated quantum benefits across diverse fields, including optimisation problems \cite{413Blekos}\cite{414Amaro}, cybersecurity applications \cite{327Raheman}\cite{415Shahwar}, and molecular sciences \cite{319Motta}\cite{321Outeiral}\cite{416Zhang}. Within this expanding quantum landscape, quantum neural networks (QNNs) represent a compelling approach that leverages parameterised quantum circuits (PQCs) \cite{33Benedetti}\cite{85Mitarai} to address the computational limitations faced by classical models, potentially offering new pathways for processing high-dimensional data \cite{417Chen}.

QNNs are capable of tackling the same learning tasks as their classical counterparts. Demonstrated use cases range across both regression tasks, such as energy output forecasting \cite{332Santos}\cite{38Hong} and temperature forecasting for infrastructure monitoring \cite{418Moon}, and classification problems including medical image analysis \cite{262Khmelnytskyi}\cite{123Landman} and natural language sentiment classification \cite{385Li}\cite{419Ouamane}. This research focuses on the application of QNNs on time series data, specifically within financial datasets. Financial datasets generate time series data with complex distributions and many underlying characteristics, often making them challenging to predict \cite{420Zhang}\cite{349Tang}. Consequently, forecasting this data has been extensively explored using classical neural networks \cite{290Azoff}\cite{289Sako}\cite{94Cao} and, more recently, using QNNs as well \cite{84Rivera-Ruiz}\cite{426Choudhary}. The QNNs developed in this work focus on forecasting the GBP/USD spot exchange rate, providing a case study to investigate the models' capabilities and assess their potential performance.

While QNNs generally consist of equivalent architectures to one another, the individual components used to build these models vary across the literature. This modularity allows for the flexible interchange of the various methods for the individual components of the model,  providing a solid foundation for comparing different methodological approaches and methods. Similar methodologies are applied by Hur et al. to discover an optimal quantum convolutional neural network (QCNN) for image classification \cite{99Hur}, and by Moussa et al. in a variety of applications \cite{335Moussa} on datasets from the OpenML-CC18 benchmark suite \cite{347Bischl}. However, these works remain confined to classification problems and do not address the unique challenges associated with regression, particularly in the context of financial time series. By incorporating a broader range of parameters in the quantum circuit, alongside a wider set of performance metrics to evaluate models, this work presents a rigorous and application-focused methodology contributing valuable insights unique to this study.

This study begins by exploring QNNs, identifying and comprehensively analysing the methods most commonly used within each architectural component. The aim is to provide a thorough understanding of the QNN architectures and their constituent individual components. Building on this theoretical foundation, these insights are applied to the task of financial time series forecasting. A systematic grid search is conducted across methodological combinations, in which individual methods for each QNN component are interchanged, resulting in 1,368 unique QNN model configurations. This comprehensive approach enables the exploration of a broad design space, facilitating a detailed evaluation of model performance across diverse architectural variations. The resulting models are analysed to uncover performance patterns, and identify effective configurations for addressing the challenges of time series forecasting. A classical BiLSTM baseline is established to contextualise QNN performance, with the best-performing QNN configurations demonstrating competitive and in several cases superior predictive accuracy. In addition, the impact of real quantum hardware noise on model performance is investigated through execution on the IQM Emerald device, providing insight into the practical hardware constraints associated with current quantum systems. An in-depth discussion of these findings is provided following the implementation of the QNN. This approach not only provides a deeper understanding of how various components interact within a QNN, contributes towards identifying an optimal configuration, and highlights the challenges associated with NISQ devices, but also provides a deeper understanding into many more complex quantum models for which QNNs serve as a foundational architecture.

The remainder of this paper is organised as follows. Section \ref{sec:LitRev} surveys the existing literature on QNN architectures and their applications. Section \ref{sec:QNN} examines the constituent components of a QNN, detailing the most commonly used methods for each stage and how they may be combined to construct the general QNN architecture. Section \ref{sec:Method} describes the experimental methodology, including the dataset, model configurations, and evaluation metrics employed in the comparative study. Section \ref{sec:Results} presents the results across all 1,368 model configurations, evaluating accuracy, computational cost, convergence behaviour, and the impact of quantum hardware noise on model accuracy through execution on the IQM Emerald device. Section \ref{sec:Discuss} interprets these findings, examining the relationships between architectural choices and model performance, and drawing broader implications for hardware-aware QNN design on near-term devices. Finally, Section \ref{sec:Conc} summarises the key findings and outlines directions for future research.

\section{Literature Review} \label{sec:LitRev}
As a flexible modelling framework, the QNN can function both as a standalone model, and as a core architectural component within more complex quantum architectures,  including  recurrent and convolutional variants. These architectures often retain common elements of QNNs, such as data encoding and the ansatz, while differing in how these elements are organised to process information. The quantum recurrent neural network (QRNN) exemplifies this approach \cite{42Bausch}\cite{14Li}\cite{97Siemaszko}, retaining unmeasured qubits between circuit repetitions to propagate a quantum hidden state analogous to that found in classical RNNs. Extending this concept further, \cite{25Chen} constructs a quantum long short-term memory (QLSTM) by embedding a quantum circuit within each gate of the classical LSTM cell structure, with \cite{198Yan} advancing this again, through a convolutional QLSTM formulation. A notably different architectural variant appears in \cite{368Cong}, where successive qubit measurements progressively reduce the quantum state dimensionality to implement convolutional-style processing entirely within the quantum circuit. 

QNNs have been applied across an equivalent range of tasks and domains as classical neural networks, as evidenced throughout recent literature. Time series forecasting represents the predominant regression application for QNNs, with the non-linear nature of these models inherently capable of capturing the complex dynamics present in sequential data \cite{235Zhang}. Financial markets provide particularly compelling use cases due to the heavy-tailed distributions and long-range temporal correlations that pose forecasting challenges \cite{420Zhang}\cite{349Tang}, which quantum approaches are uniquely suited to. Several studies have demonstrated QNN-based forecasting across financial domains, such as \cite{84Rivera-Ruiz}, which employs a hybrid architecture with classical preprocessing and postprocessing layers surrounding a central QNN to predict USD/EUR exchange rates. A decomposition-ensemble approach also using a QNN is employed in \cite{429Liu} to forecast stock prices on the Shenzhen Stock Exchange. Hybrid quantum-classical approaches for financial time-series regression are explored in \cite{426Choudhary}, and \cite{14Li} forecasts key stock market indicators using a QRNN.

Beyond financial markets, QNNs have been employed across a diverse range of further time series domains. Renewable energy forecasting has attracted particular attention, with \cite{103Ceschini} combining a QNN layer with classical LSTM components to predict solar power output, and both \cite{26Khan} and \cite{430Sagingalieva} using QLSTM architectures for the same purpose. Solar irradiance prediction is addressed in \cite{332Santos} using a standalone QNN and in \cite{196Yu} through a QLSTM formulation. Environmental and atmospheric forecasting also features prominently, with a convolutional QLSTM  modelling sea-drift trajectories governed by complex oceanographic and meteorological factors \cite{198Yan}. Wind energy applications are similarly represented, with \cite{427Hong} combining a residual LSTM with a QNN for 24-hour ahead wind speed forecasting using an offshore wind farm dataset. Physiological time series, such as patient vital sign monitoring data, have also been forecast using hybrid QNN frameworks \cite{421Iele}. Beyond physical systems, QNNs have been applied to infrastructure management through cloud computing workload prediction \cite{331Gupta}, with \cite{428Tripathi} extending this to carbon-efficient job arrival forecasting using a hybrid quantum-CNN architecture, highlighting the broad utility of quantum forecasting approaches across scientific and industrial domains.

The systematic comparison of QNN component configurations has been explored in several prior works. Hur et al. \cite{99Hur} conduct a comparative study of nine ansatz configurations within a QCNN framework for binary image classification on the MNIST and Fashion MNIST datasets, finding that ansatz with greater numbers of free parameters generally achieve higher accuracy, though at increased computational cost. Moussa et al. \cite{335Moussa} take a broader hyperparameter optimisation approach, evaluating over ten circuit design choices which include encoding strategy, entangling gate selection, circuit connectivity, and variational layer configuration. This study is conducted across low-dimensional classification datasets from the OpenML-CC18 benchmark suite \cite{347Bischl}, using functional ANOVA to rank hyperparameter importance. Their key finding is that learning rate is the most critical hyperparameter, while the choice of entangling gate has the least impact on performance. Ahmed et al. \cite{422Ahmed} extend this comparative approach to include noise robustness, evaluating multiple HQNN architectures across varying circuit configurations before assessing the best-performing models under several quantum noise channels, finding that architecture choice significantly influences noise resilience. 

Despite the systematic nature of these studies, none address regression tasks, and all are confined to classification benchmarks with limited circuit scale. This work extends this comparative methodology to financial time series forecasting, while incorporating a broader set of circuit configurations across multiple encoding methods, ansatz, qubit counts, and layer depths. Additionally, further performance dimensions including computational cost, convergence behaviour, and the impact of real quantum hardware noise on predictive accuracy are evaluated.

\section{Quantum Neural Network Architecture} \label{sec:QNN}
\subsection{Structure of a QNN} \label{subsec:StrucQNN}
QNNs adopt a multi-stage architecture that mirrors classical neural networks in their optimisation approach, while differing in how they process information. Where classical neural networks transform information through interconnected nodes and trainable weights, QNNs process information utilising quantum circuits composed of trainable parameterised gates. The interaction between classical data and quantum computation introduces additional encoding and measurement steps, contributing to a modular architecture where multiple implementation choices are available at each stage.

\begin{figure}[htbp]
    \centering
    \includegraphics[width=0.6\textwidth]{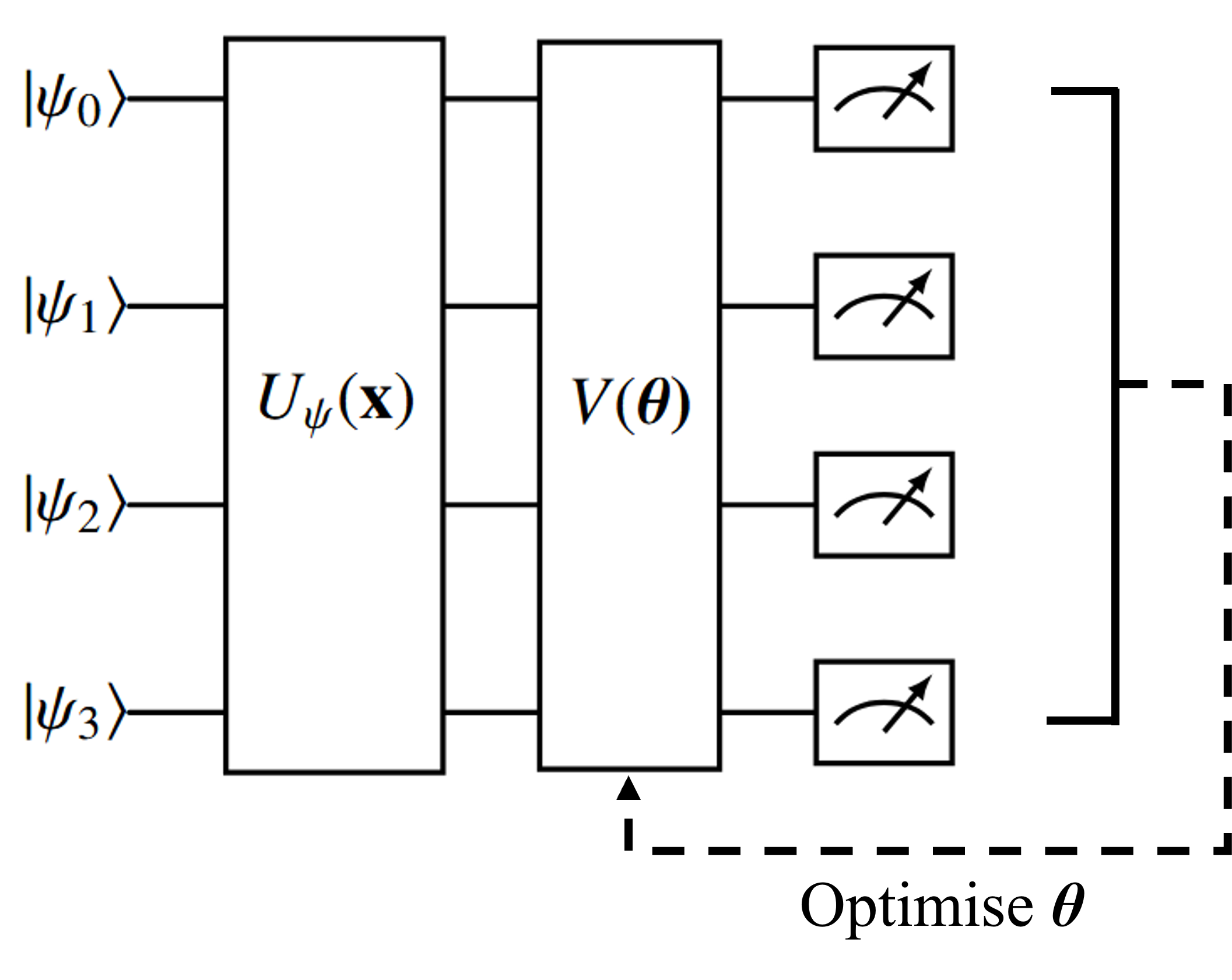}
    \caption{The quantum circuit of a basic QNN. $\ket{\psi_i}$ are the initial quantum states, $U_\psi(\textbf{x})$ is the data encoding circuit, and $V(\bm{\theta})$ is the ansatz, which is then followed by measurement on all qubits. Parameters $\bm{\theta}$ are then optimised and updated in $V(\bm{\theta})$.}
    \label{fig:basicqnn}
\end{figure}

The QNN workflow considered in this study comprises three sequential phases, each offering flexibility in method selection. (1) \textit{Data encoding} transforms classical input vectors into quantum states through various embedding schemes, making the information compatible with quantum processing. Multiple encoding strategies are available, each with distinct advantages for different data types and computational constraints. (2) The \textit{quantum processing} manipulates the encoded state through the ansatz, the variational component of the PQC, whose gate structure and connectivity pattern can be tailored to the specific learning task. The choice of ansatz significantly impacts both the model's expressiveness and computational requirements. (3) \textit{Optimisation} iteratively refines the circuit parameters to minimize prediction error, analogous to weight updates in classical networks but requiring quantum-specific measurement and gradient computation procedures. The optimisation cycle begins with quantum measurement to extract classical output from the circuit, as shown in Figure \ref{fig:basicqnn}. A cost function then compares these measured outputs to the target values to determine the discrepancy between them. The cost function gradients with respect to circuit parameters are evaluated using techniques adapted for quantum circuits. These gradients drive parameter updates via standard optimisation algorithms, with the process repeating until convergence.

The subsequent sections detail the specific methodologies evaluated in this comparative study of QNN architectures for time series forecasting. Three core components are analysed: the data encoding schemes, the ansatz designs, and the cost function formulations. Within each component, we examine multiple established approaches to assess their relative performance characteristics. Additionally, the impact of qubit count on model behavior is assessed.

\subsection{Data Encoding} \label{subsec:encoding}
Encoding classical data into quantum states is a fundamental step in quantum machine learning, as it directly affects circuit expressivity, qubit requirements, and model trainability. This study considers three widely used encoding methods:  amplitude encoding, angle encoding and circuit encoding. Each offers different trade-offs between qubit efficiency, circuit complexity, and practical implementation on near-term quantum hardware.

\subsubsection*{Amplitude Encoding}
Amplitude encoding is the first encoding method applied in this study. Under this scheme, a classical input vector $\textbf{x} = (x_1, \dots, x_n)^T \in \mathbb{R}^n$ is mapped to the probability amplitudes of a quantum system comprising $N$ qubits with $n=2^N$. The mapping is realised by a unitary transformation $U_\psi(\textbf{x})$ acting on an initial reference state, producing
\begin{equation} \label{eq:amp-enc}
\ket{\psi(\textbf{x})} = \frac{1}{\|\textbf{x}\|}\sum_{i=0}^{2^N-1}x_i\ket{i}
\end{equation}
\noindent with $\ket{i}$ denoting the $i$th computational basis state and the prefactor $1/\|\textbf{x}\|$ enforcing L2 normalisation. Amplitude encoding allows $2^N$ values to be encoded into $N$ qubits, providing a substantial data density advantage and an exponential improvement over alternative encoding schemes. This advantage comes at the cost of circuit depth, which grows as $O(\text{poly}(N))$ and can limit the method's practicality on near-term devices at higher qubit counts.

\subsubsection*{Angle Encoding}
Angle encoding \cite{70Schuld} is the second method employed in this study. Each normalised classical data point $x_i \in [0, \pi)$ is used to parameterise a single-qubit rotation, with the rotation angle determining the resulting qubit state. Applied to a single qubit, the encoding produces $\ket{\psi(x_i)} = \cos(x_i/2)\ket{0} + \sin(x_i/2)\ket{1}$. For an input vector $\textbf{x} = (x_1, \dots, x_N)^T \in \mathbb{R}^N$ across $N$ qubits, the encoded state is the tensor product of the individual qubit encodings,
\begin{equation}
    \ket{\psi(\textbf{x})} = \bigotimes^N_{i=1}\left(\cos\left(\frac{x_i}{2}\right)\ket{0} + \sin\left(\frac{x_i}{2}\right)\ket{1}\right),
\end{equation}
\noindent realised by the unitary $U_\psi(\textbf{x}) = \bigotimes^N_{i=1} R_y(x_i)$, where $R_y$ denotes the standard single-qubit rotation about the $y$-axis. This implementation is attractive because the encoding circuit has constant depth regardless of the number of qubits, keeping the computational overhead low. However, the qubit cost grows in direct proportion to the size of the input, with $N$ qubits required to encode $N$ data points, which can be restrictive on current hardware. Splitting the input across multiple circuit executions can mitigate this constraint at the cost of additional circuit runs.

\subsubsection*{Circuit Encoding}
Circuit encoding \cite{74Havlicek} is the third method applied in this study, and is intended to construct a richer feature map than the simpler encoding schemes above. Circuit encoding generates states that are believed to be classically intractable, a property that has been argued to constitute a form of quantum advantage \cite{33Benedetti, 74Havlicek}. The implementation used here follows the construction of Havlíček et al. \cite{74Havlicek}, in which the feature map is realised by the unitary $U_\psi(\textbf{x}) = V_\psi(\textbf{x})H^{\otimes N}$, with
\begin{equation} \label{eq: circ-enc}
    V_\psi(\textbf{x}) = \exp\left(i\sum_{S\subseteq[N]}\varphi_S(\textbf{x})\prod_{j\in S}Z_j\right).
\end{equation}
\noindent The sum runs over subsets $S$ of the $N$ qubits, $Z_j$ is the Pauli-Z operator acting on the $j$th qubit, and the coefficients $\varphi_S(\textbf{x})$ carry the classical input. To keep the circuit tractable, only interactions with $|S| \le 2$ are retained, restricting the implementation to single and two-qubit gates. The single-qubit coefficients are taken as $\varphi_{\{i\}}(\textbf{x}) = x_i$ and the two-qubit coefficients as $\varphi_{\{i,j\}}(\textbf{x}) = (\pi - x_i)(\pi - x_j)$, for normalised data points $x_i \in [0, \pi)$ as in angle encoding. The resulting circuit encodes $N$ data points with $N$ qubits at the same density as angle encoding, but at the cost of greater circuit depth and computational overhead, with this depth increasing further if the encoder block is repeated to enrich the feature map. When applied in the QNN for time series forecasting, circuit encoding is used with $S=2$, $\varphi_{\{i\}}(\textbf{x})=x_i$, and $\varphi_{\{i, j\}}(\textbf{x}) =  x_ix_j$. The encoding circuit is repeated 4 times to increase the complexity of the feature map further. 

\subsection{Ansatz} \label{subsec:ansatz}
Following encoding, the resulting quantum state undergoes processing through the ansatz, a variational circuit containing adjustable parameters. This circuit employs parameterised quantum gates whose parameters are tuned during training to learn the desired mapping. The architecture typically incorporates both single-qubit rotations and multi-qubit entangling operations,  both to explore the solution space and to leverage the circuit's entangling capabilities. The ansatz used in this work employ rotation gates as the parameterised gates and CNOTs as the non-parameterised entangling gates, though alternative gate sets exist \cite{97Siemaszko}.

The circuit topology critically determines QNN performance. Since the parameter space grows polynomially with system size while the full quantum state space expands exponentially \cite{33Benedetti}, practical circuits may only be able to access a limited portion of the full Hilbert space. Sim et al. \cite{75Sim} characterise an ansatz through two key measures: \textit{expressibility}, which quantifies how broadly the circuit can reach across different quantum states, and \textit{entangling capability}, measured via the Meyer-Wallach entanglement. Effective circuit design requires balancing these quantum advantages against practical constraints. Deeper, more highly connected circuits typically offer greater expressiveness and entangling power, but at the expense of increased noise sensitivity and execution time. On near-term hardware with limited coherence times and qubit connectivity, this trade-off generally favours shallower and less connected designs.

Adding multiple repetitions of the base circuit can enhance both expressiveness and entangling power \cite{75Sim}, though this comes with corresponding increases in parameter count and gate depth.

\clearpage \newpage
\begin{figure}[!h]
    \centering
    \includegraphics[width=\textwidth]{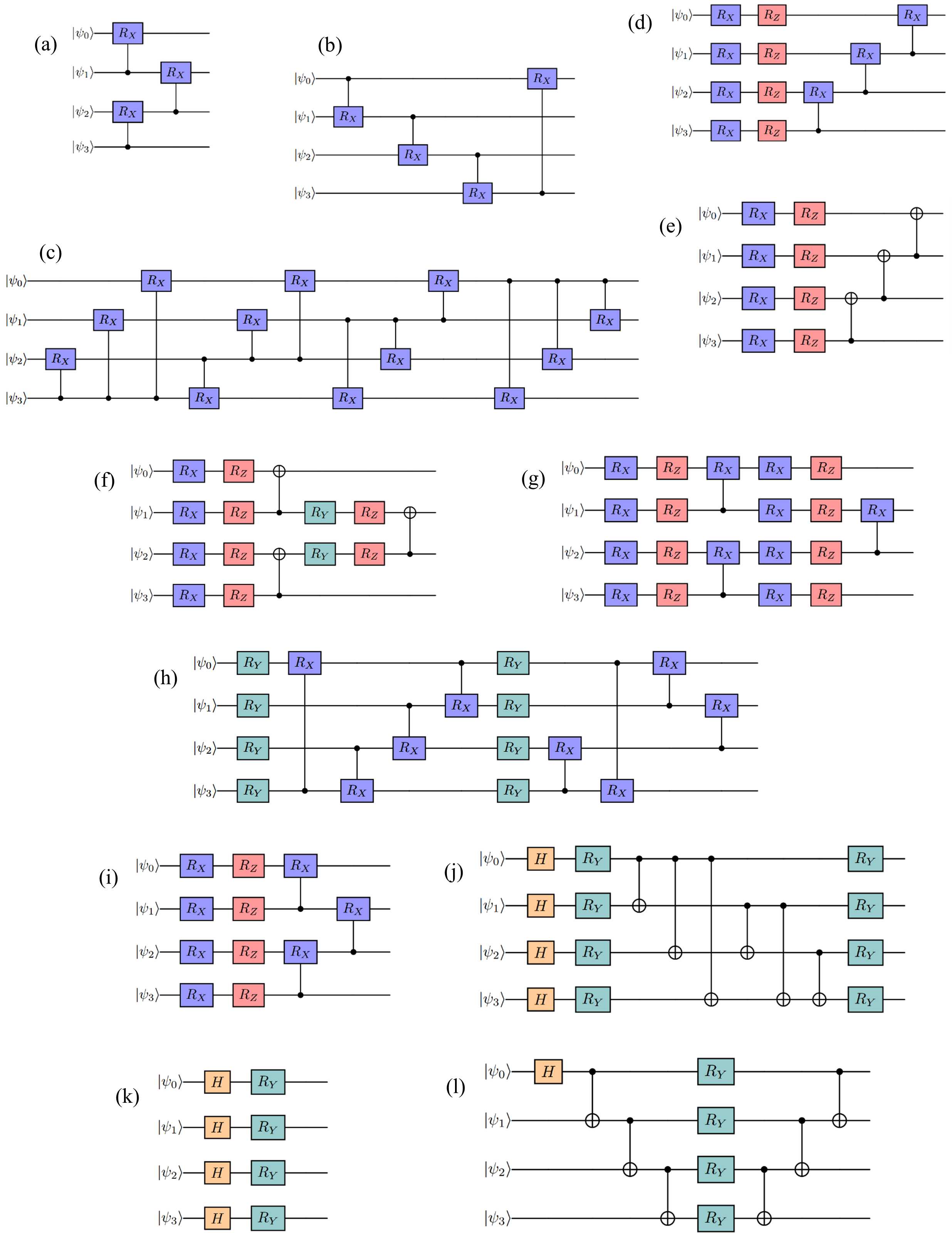}
    \caption{A collection of ansatz used in this study, each with varying expressibility, entangling capabilities and circuit depth.}
    \label{fig:ansatz}
\end{figure}
\clearpage \newpage

To assess how different circuit architectures affect QNN performance, we evaluate twelve distinct ansatz configurations drawn from the literature, as illustrated in Figure \ref{fig:ansatz}. These circuits span a wide range of expressibility, entangling capability, and computational complexity, enabling comprehensive analysis of their impact on model performance. While Figure \ref{fig:ansatz} displays each ansatz for four qubits, all designs scale to arbitrary qubit counts while preserving their structural characteristics.

The configurations (a), (b), and (c) all derive from the QRNN framework of Li et al. \cite{14Li} for temporal sequence modeling. Ansatz (a) implements a 'nearest-neighbor' connectivity pattern with minimal expressibility and entanglement but low computational overhead. Ansatz (c) uses a fully-connected topology to achieve high expressibility and strong entanglement, though this requires significantly more gates and computational resources. Ansatz (b) represents an intermediate design balancing these competing factors. Ansatz (d) and (e) originate from \cite{75Sim} which studies the expressibility and entangling capability of quantum circuits, many of which are included in the collection of ansatz here. Ansatz (f) appears in the Josephson sampling methodology \cite{244Geller}, while ansatz (g) serves as the variational encoder in the QVECTOR algorithm \cite{245Johnson}. Ansatz (h) and (i) are also both circuits from \cite{75Sim}, where ansatz (h) is inspired by the circuit-block construction in \cite{72Schuld}, and ansatz (i) is similar to the 'nearest-neighbour' configuration from ansatz (a) but with additional parameters. The remote sensing classification work in \cite{186Sebastianelli} contributes three additional designs: ansatz (j) is the 'real-amplitudes' structure with CNOT-based entanglement, ansatz (k) omits two-qubit gates entirely to eliminate entanglement, and ansatz (l) is the 'Bellman' configuration.

Since computational cost correlates directly with parameter count and circuit depth, details of how these quantities scale with the number of qubits $N$ and circuit layers $L$ for each ansatz are shown in Table \ref{tab:ansatz}.

\begin{table}[htbp]
\caption{Parameter count and circuit depth scaling for each ansatz with respect to $N$ and $L$.}\label{tab:ansatz}

\begin{tabularx}{0.8\textwidth}{ L{0.4} L{1.3} L{1.3} }
\toprule
\textbf{Ansatz} & \textbf{Number of Parameters} & \textbf{Circuit Depth} \\
\midrule
\centering a & $(N-1)L$ & $2L$ \\
\centering b & $NL$ & $NL$ \\
\centering c & $(N-1)NL$ & $(N-1)NL$ \\
\centering d & $(3N-1)L$ & $(N+1)L$ \\
\centering e & $2NL$ & $(N+1)L$ \\
\centering f & $(4N-4)L$ & $\begin{cases}
(n + 1 + \lfloor\frac{n-2}{2}\rfloor)L, & \text{if } n \leq 9 \\
(n + 4)L, & \text{if } n \geq 10
\end{cases}$ \\
\centering g & $(5N-1)L$ & $6L$ \\
\centering h & $4NL$ & $\lfloor\frac{3N+7}{2}\rfloor \cdot L$ \\
\centering i & $(3N-1)L$ & $4L$ \\
\centering j & $2NL$ & $(\frac{(N-1)N}{2}+3)L$ \\
\centering k & $NL$ & $2L$ \\
\centering l & $NL$ & $2NL$ \\
\botrule
\end{tabularx}
\end{table}

\subsection{Cost Function}
Optimisation of the QNN proceeds by adjusting the variational parameters in the ansatz to minimise a chosen cost function, which quantifies the discrepancy between the network's predictions and the target outputs. The form of the cost function depends on the task and on whether the comparison is performed in classical or quantum form.

When the QNN output is reduced to a classical data point through measurement, a standard classical cost function can be used in the same way as for classical machine learning models. For time series forecasting such as that considered in this study, the mean squared error (MSE), equivalent to an $L_2$ loss, is a common choice:
\begin{equation}
    C(\bm{\theta}) = \frac{1}{2n}\sum_{i=1}^{n}\bigl(\bm{y}_{\text{out}}(\bm{\theta}) - \bm{y}_{\text{data}}\bigr)^2,
\end{equation}
\noindent where $\bm{\theta} = (\theta_1, \dots, \theta_N)$ denotes the variational parameters being optimised, $\bm{y}_{\text{out}}$ and $\bm{y}_{\text{data}}$ are the predicted and observed values, and the sum runs over all $n$ data samples.

A quantum cost function offers an alternative formulation in which the comparison is made directly between quantum states rather than after measurement. The fidelity, which produces values in $[0,1]$, quantifies the similarity of two quantum states and provides a natural basis for such a cost function \cite{56Beer}. For pure states, comparing the QNN output taken before measurement $\ket{\psi_{\text{out}}(\bm{\theta})}$ against the encoded target $\ket{\phi_{\text{data}}}$, the cost is
\begin{equation} \label{eq: quant-cost}
    C(\bm{\theta}) = \frac{1}{n}\sum_{x=1}^n\left(1- |\braket{\phi_{data}|\psi_{out}(\bm{\theta})}|^2\right)
\end{equation}
\noindent Since fidelity is maximised at perfect agreement, the negative sign inverts this such that the optimiser minimises the cost whilst maximising the overlap between the predicted and target states.

In the application of the QNN for time series forecasting, the classical cost function uses the MSE between the circuit's output and the true data in the same manner as is seen in classical models. The output of the circuit is produced through quantum measurement of the expectation value on the Pauli-Z observable. This measurement is taken only on the first qubit, a step taken to avoid barren plateaus that is also seen in \cite{14Li}. The quantum cost function measures the quantum fidelity between the circuit's terminal quantum state and the true data. To achieve a quantum representation of the true data, the data is encoded into the first entry of the density matrix of a qubit using a quantum circuit, and so the fidelity can then be calculated. For the quantum cost function, the output of the circuit is taken as the real part of the first entry of the density matrix of the first qubit.

\section{Methodology} \label{sec:Method}
\subsection{Data} \label{subsec:data}
Financial time series have been widely used to evaluate classical neural networks \cite{274Kaastra}\cite{275Adhikari}\cite{276Cheng}, and more recently QNN-based approaches \cite{89Paquet}\cite{83Emmanoulopoulos}\cite{148Lindsay}. This study considers the GBP/USD spot exchange rate as a financial regression case study. GBP/USD is the third most actively traded currency pair in the global foreign exchange market \cite{286Cespa}, underscoring the value of accurate price prediction.

\begin{figure}[htbp]
    \centering
    \includegraphics[width=\textwidth]{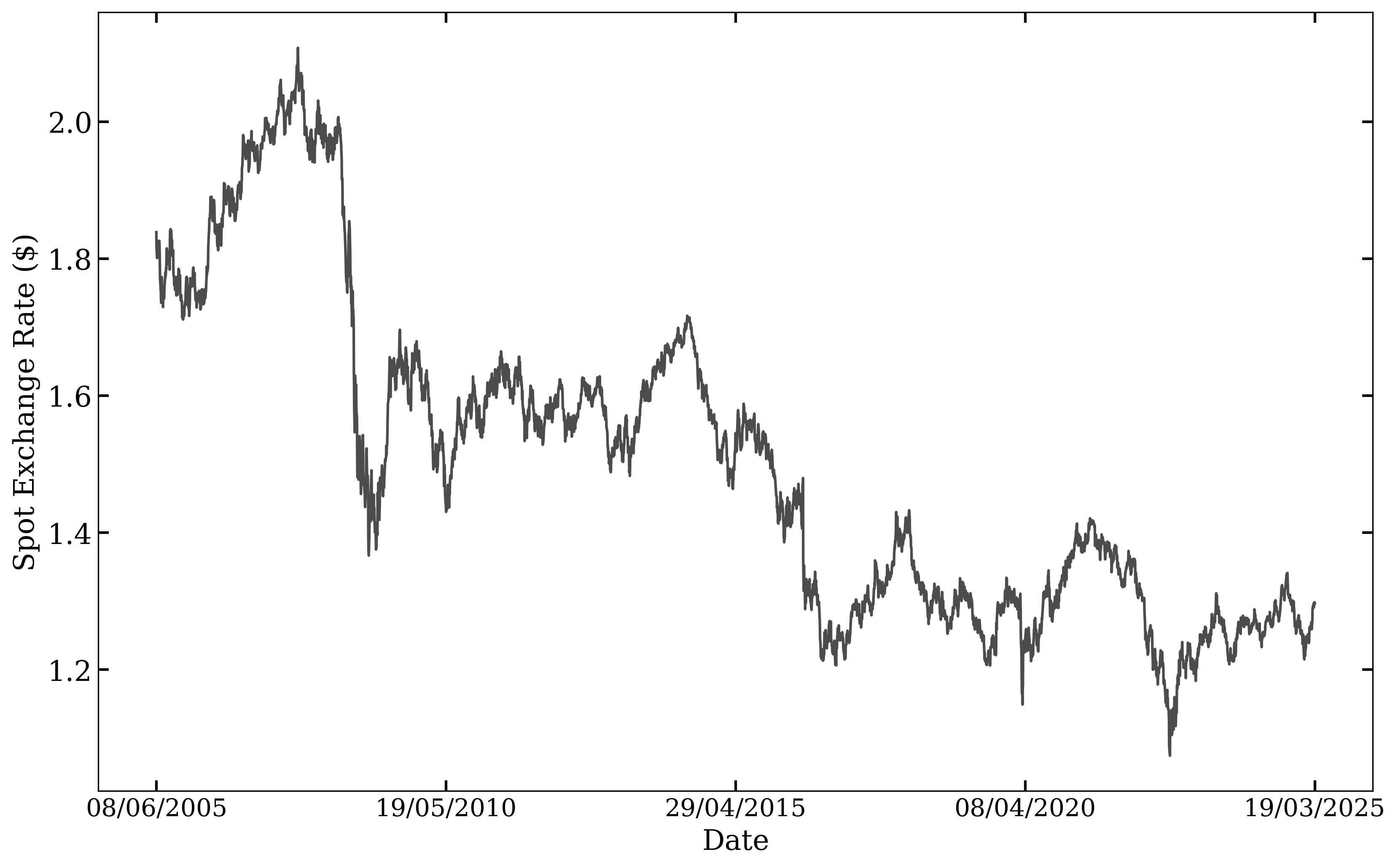}
    \caption{The spot exchange rate (\$) for the GBP/USD currency pair from 08/06/2005 to 19/03/2025.}
    \label{fig:GBPUSDdata}
\end{figure}

The spot exchange rate, defined as the price at which a currency can be exchanged at a given moment, is recorded daily at 9:30am in this dataset and sourced from the Bank of England database \cite{301Bankengland}. The dataset comprises daily spot exchange rates from 8th June 2005 to 19th March 2025, consisting of 5,000 samples ranging from \$1.0745 to \$2.1081, with an average daily movement of \$0.0066 (standard deviation \$0.0064), as shown in Figure \ref{fig:GBPUSDdata}. 

The data is split chronologically into training (3,000 samples, 08/06/2005-19/04/2017), validation (500 samples, 20/04/2017-09/04/2019), and test (1,500 samples, 10/04/2019-19/03/2025) sets in a 60/10/30 ratio. The validation set optimises hyperparameters on a subset of QNN models before final evaluation on the unseen test set.

\subsection{Experimental Design} \label{subsec:exp-des}
To evaluate the optimal configuration for QNN-based time series forecasting, this study conducts a systematic grid search across all methodological combinations described in Section \ref{sec:QNN}, applied to the GBP/USD spot exchange rate. This comprehensive approach examines how individual QNN components interact with one another, providing insights into their combined effects on forecasting performance. The implementation uses PennyLane \cite{239PennyLane} for its well-established quantum machine learning capabilities.

The experimental design spans five configuration dimensions. Qubit counts of 4, 6, 8, 10, and 12 are evaluated, with hardware limitations preventing testing beyond 12 qubits. All three encoding methods (amplitude, angle, and circuit) are assessed alongside all twelve ansatz configurations illustrated in Figure \ref{fig:ansatz}. The ansatz are repeated in layers of counts 1, 2, 4, and 8, which is reduced to 1, 2, and 4 for 12 qubit models due to computational constraints. Finally, both classical MSE and quantum fidelity cost functions are included. The circuit output is obtained via a Pauli-Z measurement on the first qubit for the classical cost function, and the real part of the first density matrix entry of the first qubit for the quantum cost function. From here, gradients are computed using the backpropagation algorithm, an approach made possible by the classical simulation environment, as real quantum hardware would require quantum-compatible methods such as the parameter-shift rule. Parameter updates are then handled by the Adam optimiser.

\hfill \break
\begin{center} 
\begin{adjustbox}{width=1\textwidth}
\begin{quantikz}
    \ket{0}   &  \gate{H} \gategroup[4,steps=2,style={dashed,rounded corners, fill=red!20, inner xsep=2pt},background, label style={label position=below,anchor=north,yshift=-0.2cm}]{Data Encoding} & \gate{R_Y}  &  \gategroup[4,steps=14,style={dashed,rounded corners, fill=blue!20, inner xsep=2pt},background, label style={label position=below,anchor=north,yshift=-0.2cm}]{Parameterised Quantum Circuit}         &           & \gate{R_X}  &          &           & \gate{R_X}  &          &          &  & \gate{R_X}  & & \ctrl{3} & \ctrl{2} & \ctrl{1} & \\
    \ket{0} & \gate{H} & \gate{R_Y}  &           & \gate{R_X}  &           &          & \gate{R_X}  &           & \ctrl{2} & \ctrl{1} &  &\ctrl{-1} &     &    &          & \gate{R_X} & \\
    \ket{0} & \gate{H} & \gate{R_Y}  & \gate{R_X}  &           &           & \ctrl{1} & \ctrl{-1} & \ctrl{-2} &          & \gate{R_X} &      &       &     &     & \gate{R_X} &          & \\
    \ket{0} & \gate{H} & \gate{R_Y}  & \ctrl{-1} & \ctrl{-2} & \ctrl{-3} & \gate{R_X} &           &           & \gate{R_X} &          &       &      &  & \gate{R_X} &          &          & 
\end{quantikz}
\end{adjustbox}
    \captionof{figure}{A representative quantum circuit implemented in the QNN for 4 qubits, combining angle encoding (red box) with a single layer of ansatz (c) (blue box).}
    \label{fig:modelqcirc}
\end{center}

A non-overlapping window approach is used to structure the input data, with window sizes determined by the encoding method and qubit count. For amplitude encoding with $N \leq 5$, the data is partitioned into non-overlapping windows of $2^N$ data points. For $N > 5$, windows are fixed at 32 points encoded across 5 qubits, with remaining qubits reserved for state manipulation. Angle and circuit encoding partition the data into non-overlapping windows of $N$ data points. 

Feature scaling is applied independently for each encoding method. Amplitude encoding applies $L^2$ normalization to the input data groups, with output values scaled to a comparable magnitude using the corresponding group norms. For angle encoding input data is scaled to $[0, \pi/2]$, while output data is independently scaled to $[0.25, 0.75]$. For circuit encoding, both input and output data are scaled to $[0.25, 0.75]$ using a min-max scaler. The output range of $[0.25, 0.75]$ is chosen across both methods to accommodate the $[-1, 1]$ output range of the Pauli-Z measurement while permitting extrapolation beyond the training bounds in either direction, which is appropriate given that the exchange rate is strictly positive.

Training employs a learning rate of 0.1 over 300 epochs. Each configuration is executed five times with randomized initial parameters to enable statistical comparison, with results averaged across runs. All simulated experiments are conducted on an Intel Core i7-4790 CPU @ 3.60GHz with 32GB RAM.

Performance is evaluated across three metrics; prediction accuracy, computational cost, and optimisation efficiency. Prediction accuracy is assessed using mean squared error (MSE) and coefficient of determination ($R^2$), the computational cost is measured through the training time, and the optimisation efficiency is quantified by the convergence epoch, defined as the point at which the five-epoch moving average cost change falls below 0.5\%. For true data $\textbf{y} = (y_1,y_2, \dots, y_n)$ and predictions $\hat{\textbf{y}} = (\hat{y}_1, \hat{y}_2, \dots, \hat{y}_n)$:

\begin{align}
    &MSE = \frac{1}{n}\sum^n_{i=1}(y_i - \hat{y}_i)^2 \\
    &R^2 = 1 - \frac{\sum^n_{i=1}(y_i - \hat{y}_i)^2}{\sum^n_{i=1}(y_i - \Bar{y})^2}
\end{align}

\noindent where $\Bar{y}$ represents the mean of the true data.

\subsection{Classical Baseline} \label{subsec:classical}
To contextualise the performance of the QNN models, a classical BiLSTM baseline is established under two configurations. The first is a constrained model, designed to match the data conditions of the QNN study using a window size of 8 data points to emulate the 8 qubit models for angle encoding, with non-overlapping windows. To avoid overfitting under this restricted input, the constrained model employs 2 bidirectional LSTM layers with 16 units each and no convolutional preprocessing. The second is an unconstrained model, giving access to a longer lookback window of 50 data points and fully overlapping windows, with a more expressive architecture comprising a Conv1D preprocessing layer followed by 3 bidirectional LSTM layers with 256 units each. Both models use the Adam optimiser with a Huber loss function and early stopping, the same 60/10/30 train/validation/test split, and the optimal configuration for each model is selected via performance on the validation set. Input and output data are feature-scaled to $[0, 1]$ for both models. The best-performing runs of the constrained and unconstrained models achieve test $R^2$ scores of 0.9529 and 0.9805, and MSE values of $1.86 \times 10^{-4}$ and $7.5 \times 10^{-5}$ respectively, across five independent initialisations.

\subsection{Quantum Hardware Implementation} \label{subsec:hardware}
\begin{figure}[!h]
    \centering
    \includegraphics[width=\textwidth]{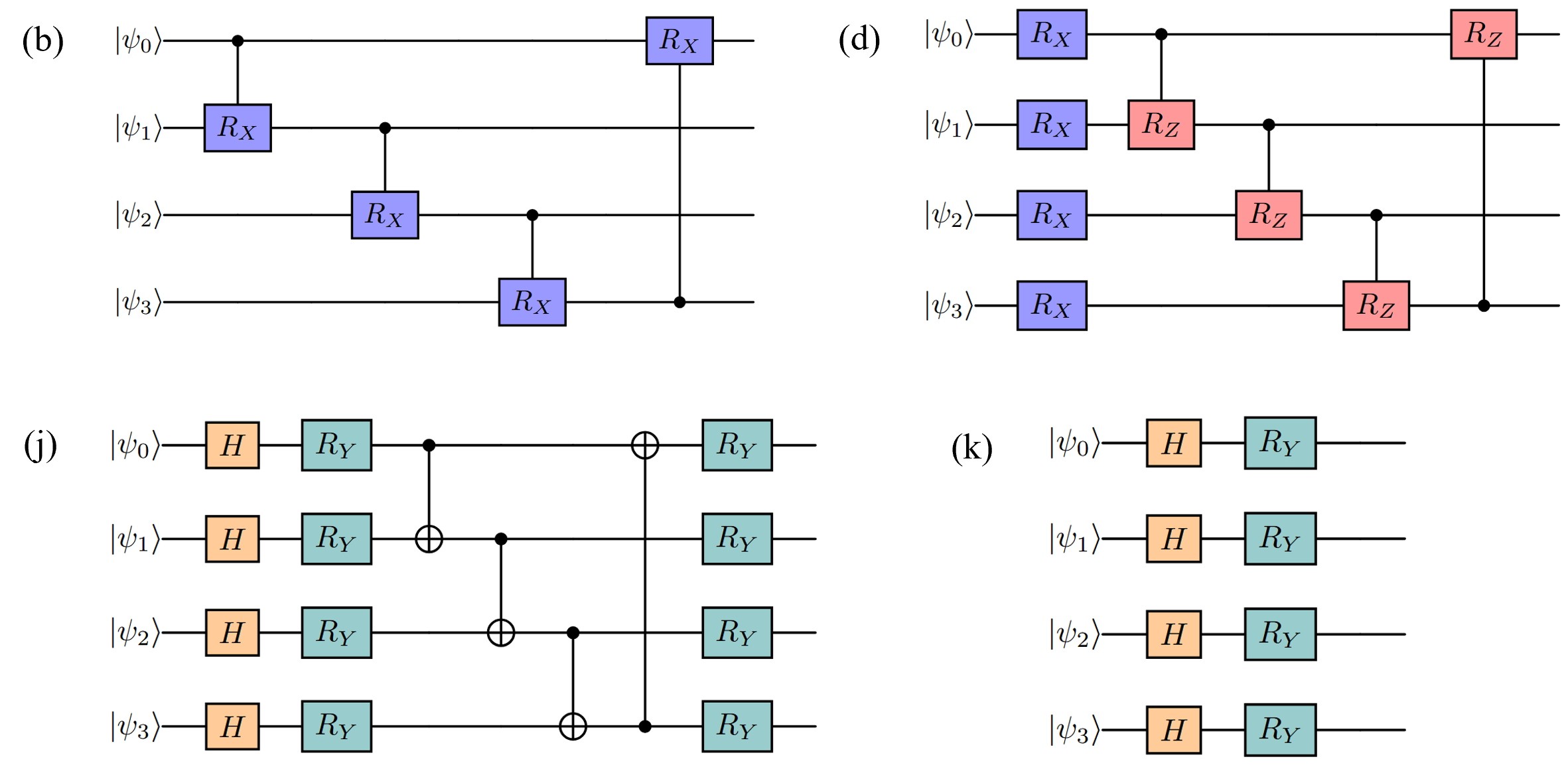}
    \caption{Modified hardware ansatz variants with two-qubit gates restricted to nearest-neighbour qubit connectivity, derived from ansatz (b), (d), (j), and (k).}
    \label{fig:ansatz-hardware}
\end{figure}

To quantify the impact of quantum hardware noise on circuit output and prediction accuracy, a subset of the grid search configurations are evaluated on both a noisy simulator and a real quantum device that replace the classical simulator during model inference. The experiment assesses three ansatz variants across qubit counts of 4, 6, 8, and 10, with layer counts of 1, 2, 4, and 8, using angle encoding and a classical MSE cost function. Angle encoding is chosen as it is the most readily implementable encoding method for near-term quantum hardware, as each feature is encoded on a single qubit and only two additional gates per qubit are required. The ansatz evaluated are modified versions of ansatz (b), (j), and (k), adapted to restrict two-qubit gates to nearest-neighbour connections in line with the qubit connectivity map of the target hardware, avoiding additional swap gates during circuit compilation. These ansatz were selected to span a range of gate types and circuit depths whilst remaining within the constraints of near-term hardware. For ansatz (k), only layer counts of 1, 2, and 4 are evaluated (as shown in Figure \ref{fig:emerald-hard}), as the absence of two-qubit gates in this ansatz results in substantially lower noise sensitivity, rendering the 8-layer configuration unnecessary for the purpose of this study. These modified circuits are illustrated in Figure \ref{fig:ansatz-hardware}.

Two execution environments are employed to allow decomposition of the hardware noise contribution. The Amazon Braket SV1 state vector simulator models readout noise only, providing a controlled baseline that enables the measurement error contribution to be isolated from other hardware noise. The IQM Emerald quantum device \cite{423IQM} introduces the full noise profile of real quantum hardware, including gate errors, decoherence, and readout noise. Comparing results across the noiseless simulator, SV1, and IQM Emerald allows for a thorough analysis of how model performance is affected by noise, along with the isolation of readout noise from the total noise contribution. All executions use 1,024 shots per circuit.

To maximise the effective shot count without additional hardware cost, Pauli-Z expectation value measurements are taken across the first four qubits and averaged to produce the circuit output. As the quantum hardware is used for inference only, all training is performed on the classical simulator with the trained parameters subsequently transferred to the hardware for evaluation. All other experimental parameters follow those described in Section \ref{subsec:exp-des}, with the exception that output data is scaled to $[0, 1]$ rather than $[0.25, 0.75]$, to avoid amplifying readout errors through the output scaling process.

An additional modified variant of ansatz (d) was also evaluated, however the use of non-native gates necessitated gate decomposition during compilation that resulted in circuit depths that exceeded the capacity of the IQM Emerald circuit compiler beyond 4 layers. The broader implications of this for hardware ansatz selection are discussed further in Section \ref{sec:Discuss}.

\section{Results} \label{sec:Results}
This section first presents the results generated from the variously configured QNNs using noiseless simulation. Following this, the hardware noise and real-device results are presented separately. For all mean results, the standard error of the mean is used as the error, calculated as $SEM = \sigma/\sqrt{n}$ where $n$ represents the number of unique models for that particular methodological variation and $\sigma$ is the standard deviation of the $n$ values. Results for unique models are averaged across the 5 runs before  mean results and errors are calculated, with the exception of the hardware noise results that are based on single runs.

\subsection{QNN Configuration Performance in Simulation}
\subsubsection*{Model Complexity}
Tables \ref{tab:enc-results} and \ref{tab:ans-results} present the mean performance across all components and runs, alongside the accuracy metrics for the best-performing individual models for the encoding methods and various ansatz respectively. All accuracy results are evaluated on the test dataset. 

\begin{table}[htbp]
\caption{Mean and best individual MSE and $R^2$ for each encoding method.}\label{tab:enc-results}
\begin{tabularx}{\textwidth}{ L{0.6} L{1.4} L{1.3} L{1.1} L{0.6} }
\toprule
Encoding & Mean MSE & Mean $R^2$ & Best MSE & Best $R^2$ \\
\midrule
Amplitude & $\textbf{0.01145} \bm{\pm} \textbf{0.00141}$ & $\bm{-}\textbf{2.168} \bm{\pm} \textbf{0.395}$ & $5.9698 \times 10^{-5}$ & 0.98481 \\
Angle & $0.08482 \pm 0.00939$ & $-21.156 \pm 2.452$ & $\textbf{5.6693} \bm{\times} \textbf{10}^{\bm{-}\textbf{5}}$ & \textbf{0.98495} \\
Circuit & $0.02969 \pm 0.00306$ & $-6.714 \pm 0.790$ & $6.3082 \times 10^{-5}$ & 0.98326 \\
\botrule
\end{tabularx}
\end{table}

\begin{table}[htbp]
\caption{Mean and best individual MSE and $R^2$ for each ansatz.}\label{tab:ans-results}
\begin{tabularx}{\textwidth}{ L{0.5} L{1.4} L{1.1} L{1.2} L{0.8} }
\toprule
Ansatz & Mean MSE & Mean $R^2$ & Best MSE & Best $R^2$ \\
\midrule
a & $0.08293 \pm 0.00777$ & $-20.965 \pm 2.045$ & $22.8673 \times 10^{-5}$ & 0.93931 \\
b & $0.01059 \pm 0.00194$ & $-1.778 \pm 0.508$ & $8.2398 \times 10^{-5}$ & 0.97854 \\
c & $0.01154 \pm 0.00288$ & $-2.067 \pm 0.770$ & $8.2204 \times 10^{-5}$ & 0.97896 \\
d & $0.00478 \pm 0.00147$ & $-0.252 \pm 0.385$ & $6.8010 \times 10^{-5}$ & 0.98270 \\
e & $0.04585 \pm 0.00974$ & $-11.117 \pm 2.558$ & $6.3082 \times 10^{-5}$ & 0.98326 \\
f & $0.03037 \pm 0.00601$ & $-7.036 \pm 1.586$ & $5.9698 \times 10^{-5}$ & 0.98481 \\
g & $0.00935 \pm 0.00216$ & $-1.442 \pm 0.566$ & $7.5152 \times 10^{-5}$ & 0.98088 \\
h & $\textbf{0.00172} \bm{\pm} \textbf{0.00063}$ & $\textbf{0.548} \bm{\pm} \textbf{0.164}$ & $6.4511 \times 10^{-5}$ & 0.98294 \\
i & $0.01001 \pm 0.00219$ & $-1.615 \pm 0.570$ & $7.2498 \times 10^{-5}$ & 0.98155 \\
j & $0.00989 \pm 0.00238$ & $-1.575 \pm 0.619$ & $\textbf{5.6693} \bm{\times} \textbf{10}^{\bm{-}\textbf{5}}$ & \textbf{0.98495} \\
k & $0.22363 \pm 0.03043$ & $-57.365 \pm 7.934$ & $42.0279 \times 10^{-5}$ & 0.88884 \\
l & $0.06315 \pm 0.01324$ & $-15.489 \pm 3.459$ & $7.0133 \times 10^{-5}$ & 0.98139 \\
\botrule
\end{tabularx}
\end{table}

Amplitude encoding achieves the best mean performance across encoding methods, followed by circuit and then angle encoding. Angle encoding can nevertheless produce excellent individual models, but appears to be more sensitive to the other configuration choices, as many angle encoding configurations perform poorly enough to reduce the mean $R^2$ to $-$21.156. For best individual models, angle and amplitude encoding achieve comparable top accuracies, with circuit encoding marginally behind.

\begin{table}[htbp]
\caption{Ansatz rankings by mean and best-individual accuracy, prioritising $R^2$ above MSE.}\label{tab:ans-rank}
\newcolumntype{Y}{>{\centering\arraybackslash}X}
\begin{tabularx}{\textwidth}{ c Y Y }
\toprule
Ranking & Mean Accuracy Metrics (Ansatz) & Best Accuracy Metrics (Ansatz) \\
\midrule
1 & h & j \\
2 & d & f \\
3 & g & e \\
4 & j & h \\
5 & i & d \\
6 & b & i \\
7 & c & l \\
8 & f & g \\
9 & e & c \\
10 & l & b \\
11 & a & a \\
12 & k & k \\
\botrule
\end{tabularx}
\end{table}

Table \ref{tab:ans-rank} ranks the performance of the various ansatz for both mean and best accuracies. Ansatz (h) achieves the greatest mean accuracy across all models, being the only ansatz to average a positive $R^2$ score, while ansatz (k) is the worst performer by a considerable margin on both metrics. The top five ansatz by best individual accuracy — (d), (e), (f), (h), and (j) — all achieve very similar high $R^2$ scores.

The $R^2$ score is bounded above by one but unbounded below, such that $R^2 \in (-\infty, 1]$. Consequently, poorly performing or failed models can produce large negative values that dominate the mean, as demonstrated by the worst performing model which achieved an $R^2$ of $-$393.544. For the remainder of this section, all negative $R^2$ scores are set to zero in visualisations so that graphs focus on models that successfully converged.

\begin{figure}[htbp]
    \centering
    \includegraphics[width=\textwidth]{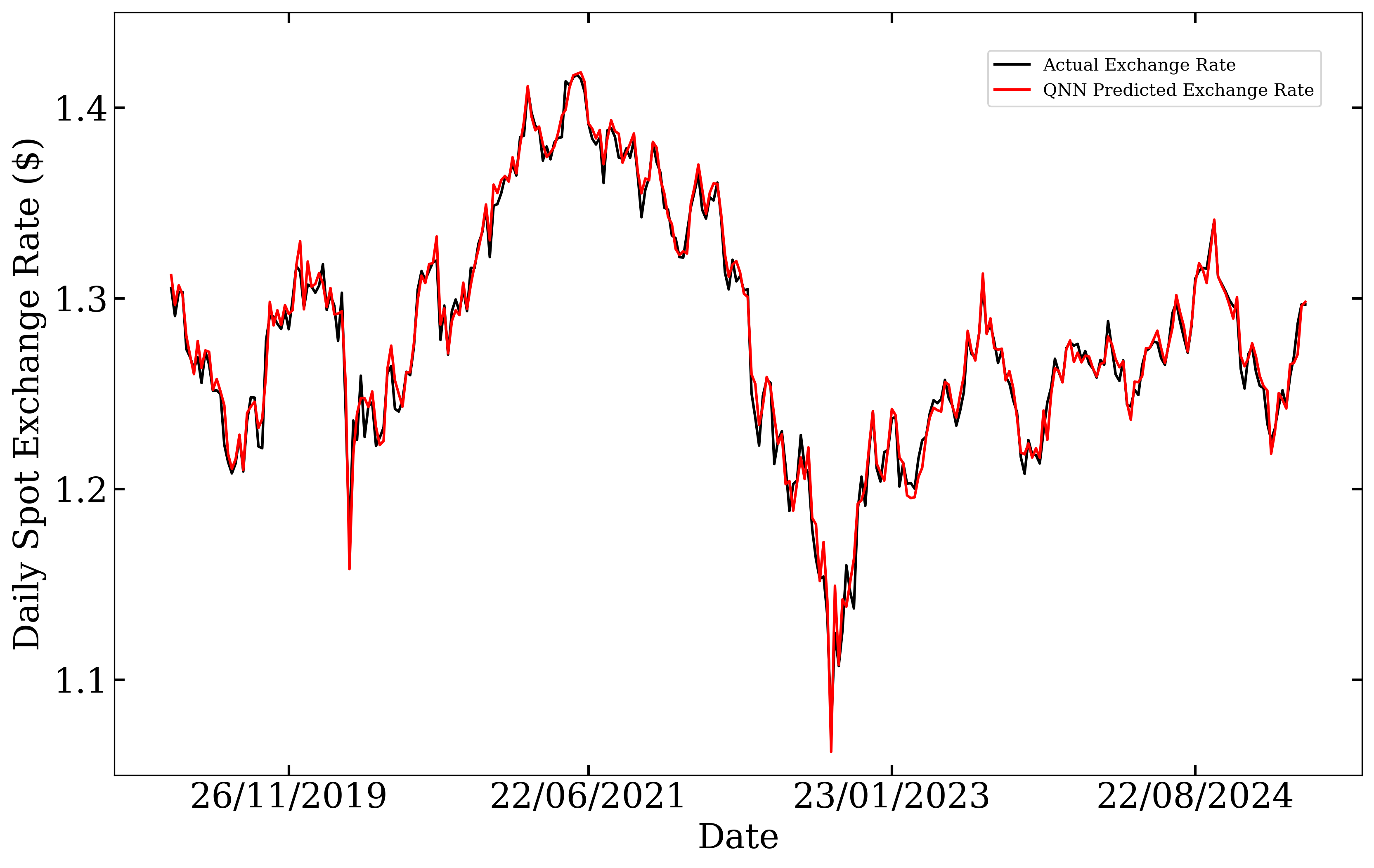}
    \caption{Best individual result for angle encoding, with an $R^2$ score of 0.98495. The model used angle encoding, ansatz (j), 4 qubits, 8 layers, a classical MSE cost function and occurred on the 5th run. The black line shows the true GBP/USD spot exchange rate and the red line shows the models simulated predicted price.}
    \label{fig:best-ang}
\end{figure}

The best individual model achieved an $R^2$ score of 0.98495 using angle encoding with ansatz (j), 4 qubits, 8 layers, and a classical MSE cost function, as shown in Figure \ref{fig:best-ang}. The best models for amplitude and circuit encoding achieved comparable results with $R^2$ scores of 0.98481 and 0.98326 respectively, producing visually similar forecasts. All three encoding methods therefore produce best individual models that exceed both the constrained BiLSTM baseline ($R^2 = 0.9529$, $MSE = 1.86 \times 10^{-4}$) and the unconstrained BiLSTM baseline ($R^2 = 0.9805$, $MSE = 7.5 \times 10^{-5}$), with the best quantum model achieving an MSE of $5.6693 \times 10^{-5}$.

\begin{sidewaysfigure}[htbp]
    \centering
    \includegraphics[width=\textwidth]{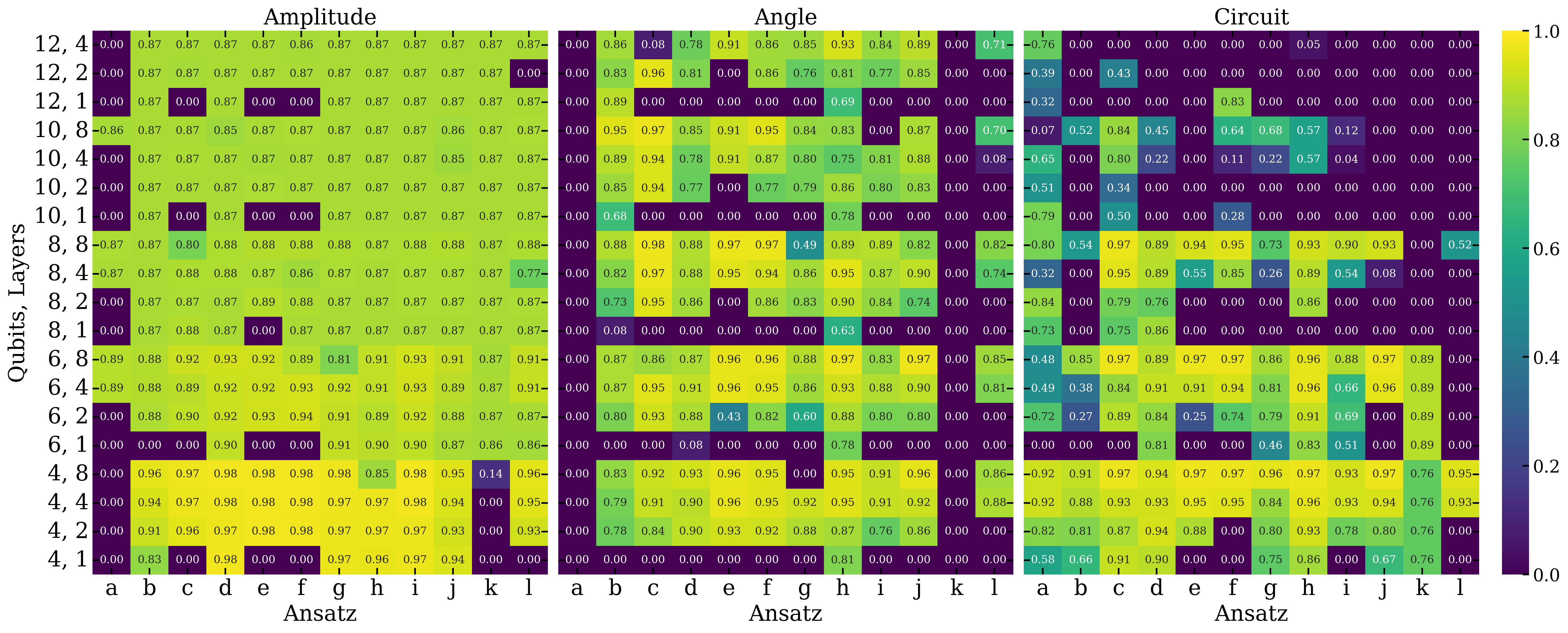}
    \caption{A heatmap showing the mean $R^2$ score across the 5 runs for the various ansatz, qubits and layers for models using a classical MSE cost function, split across the 3 encoding methods.}
    \label{fig:heat_clas}
\end{sidewaysfigure}

\begin{sidewaysfigure}[htbp]
    \centering
    \includegraphics[width=1\textwidth]{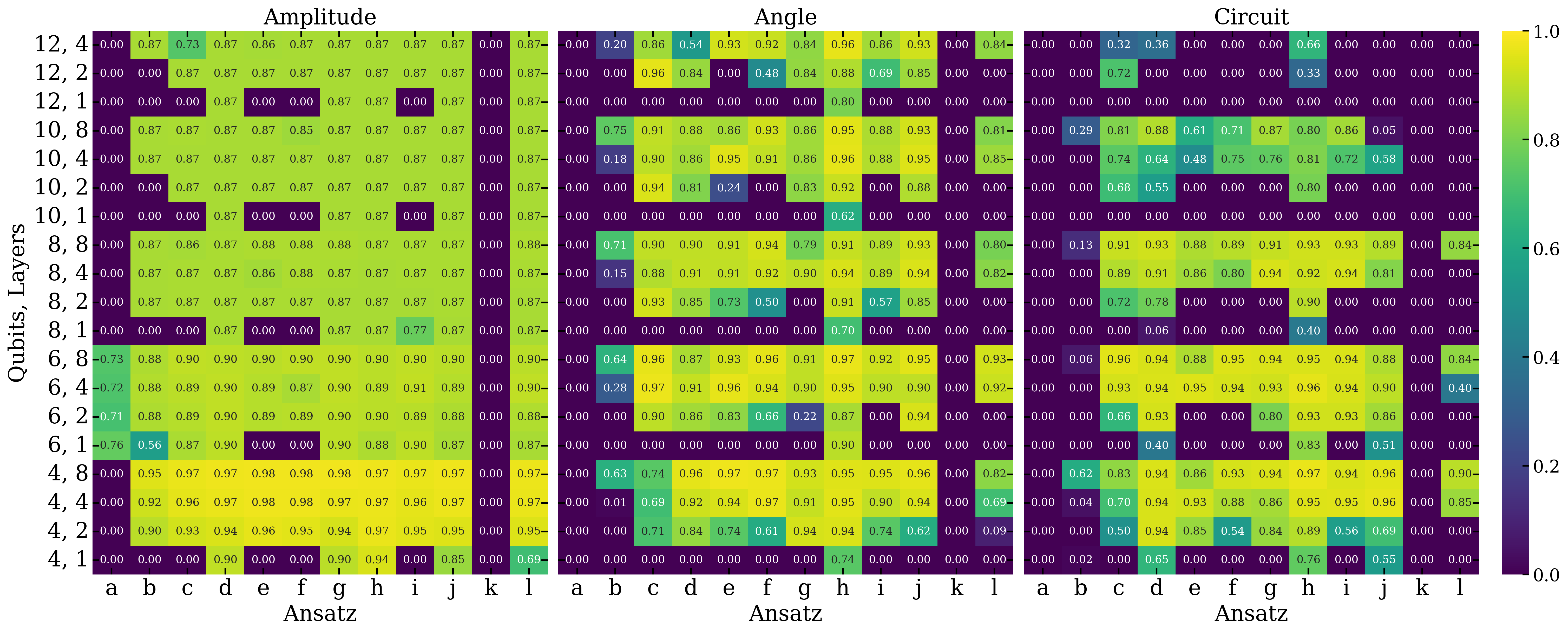}
    \caption{A heatmap showing the mean $R^2$ score across the 5 runs for the various ansatz, qubits and layers for models using a quantum fidelity cost function, split across the 3 encoding methods.}
    \label{fig:heat_quant}
\end{sidewaysfigure}

Heatmaps displaying the full set of mean $R^2$ scores across the five runs are shown in Figures \ref{fig:heat_clas} and \ref{fig:heat_quant}, for classical MSE and quantum fidelity cost functions respectively, split across the three encoding methods.

The heatmaps show that amplitude encoding converges to a solution much more frequently than angle or circuit encoding. Angle and circuit encoding appear more comparable in the heatmaps than in Table \ref{tab:enc-results}, though this stems from the negative score capping applied here.

The heatmaps also reflect the ansatz accuracy statistics. Ansatz (h) performs well across almost all models, while ansatz (a) and (k) frequently fail to converge, consistent with their position as the worst performers. However, under a classical cost function, both (a) and (k) can converge when amplitude or circuit encoding is used, but not with angle encoding. This pattern is not observed under the quantum fidelity cost function.

The effect of qubit count on accuracy varies by encoding method. Alongside influencing parameter count, two-qubit gate count, and Hilbert space dimensionality, the qubit count also slightly alters the sliding window length used for prediction (see Section \ref{subsec:exp-des}). Amplitude encoding performs best at four qubits, with accuracy decreasing at a diminishing rate up to around eight qubits before plateauing, potentially due to the decreasing fraction of qubits carrying the encoded data as qubit count increases. Angle encoding shows no strong correlation, though the best models peak at eight qubits for classical and six for quantum cost functions before declining. For circuit encoding, convergence rates and accuracy both decline as qubits increase, with best performance at four qubits across both cost functions. 

\begin{figure}[htbp]
    \centering
    \includegraphics[width=\textwidth]{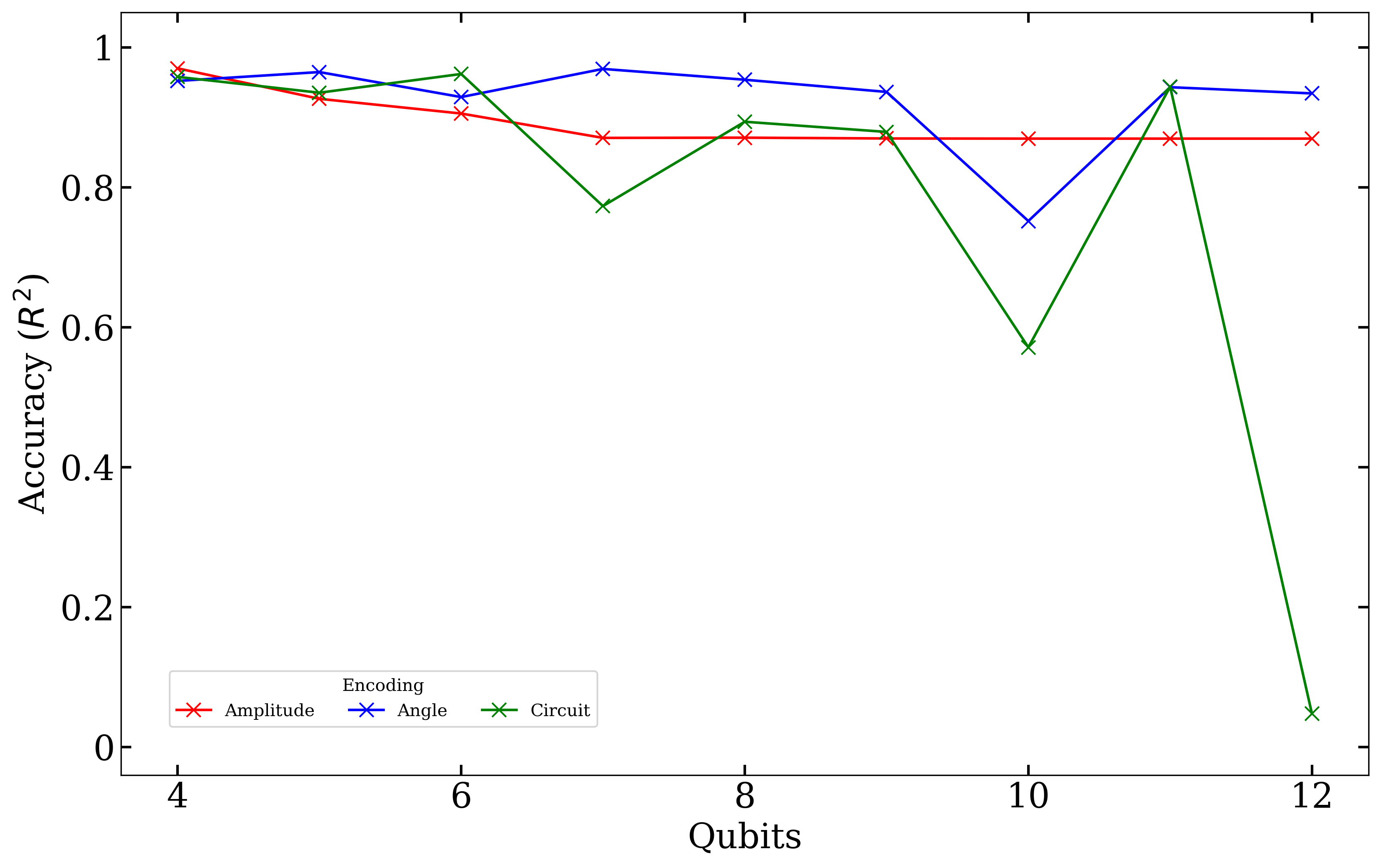}
    \caption{Mean $R^2$ score across the 5 runs for all qubits in the range $[4,12]$ for the 3 encoding methods. All models evaluated used ansatz (h) with 4 layers and a classical MSE cost function.}
    \label{fig:acc-qubits}
\end{figure}

Ansatz (h), the best-performing ansatz on average is further examined across the full range of qubit counts for each encoding method. Figure \ref{fig:acc-qubits} illustrates these comparisons using 4 layers and a classical MSE cost function, allowing the influence of qubit count and encoding choice on the performance of ansatz (h) to be assessed in greater detail.

\begin{figure}[htbp]
    \centering
    \includegraphics[width=1\textwidth]{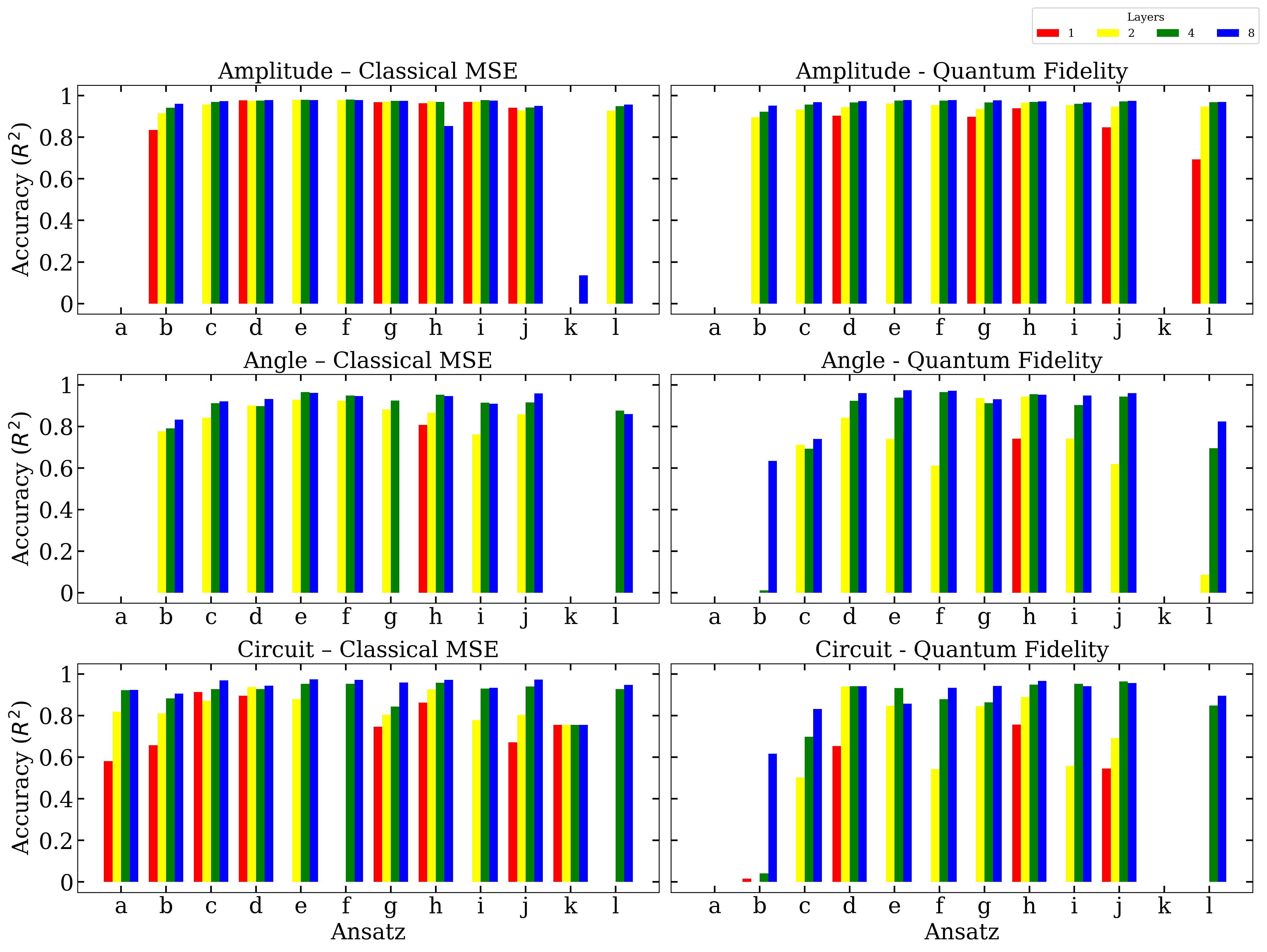}
    \caption{Mean $R^2$ score across the 5 runs for the various ansatz using a different number of layers. Results are split across the encoding methods and cost functions, and all models used 4 qubits.}
    \label{fig:acc-layers}
\end{figure}

Increasing the number of layers generally improves accuracy across all ansatz variants, encoding methods, qubit counts, and cost functions, though the magnitude of this effect varies between configurations. Figure \ref{fig:acc-layers} shows this trend for four-qubit models across encoding methods and cost functions. As mentioned earlier, negative $R^2$ values are set to zero for ease of visualisation. Consequently, missing bars, particularly for ansatz (a) and (k), indicate configurations with negative mean $R^2$ values that have been clipped to zero, showing that these ansatz variants are among the worst performers across several configurations.

Referring to Figures \ref{fig:heat_clas} and \ref{fig:heat_quant}, single-layered ansatz rarely converge under angle encoding, with the same trend observed under circuit encoding, though with more exceptions. For circuit encoding, this effect becomes more pronounced at higher qubit counts, with the best results at high qubit counts occurring only at the highest layer counts. Amplitude encoding shows similar but weaker trends; single-layer models converge more frequently, though less so under a quantum cost function, and the benefit of increasing layer count diminishes at higher qubit counts.

\begin{figure}[htbp]
    \centering
    \includegraphics[width=1\textwidth]{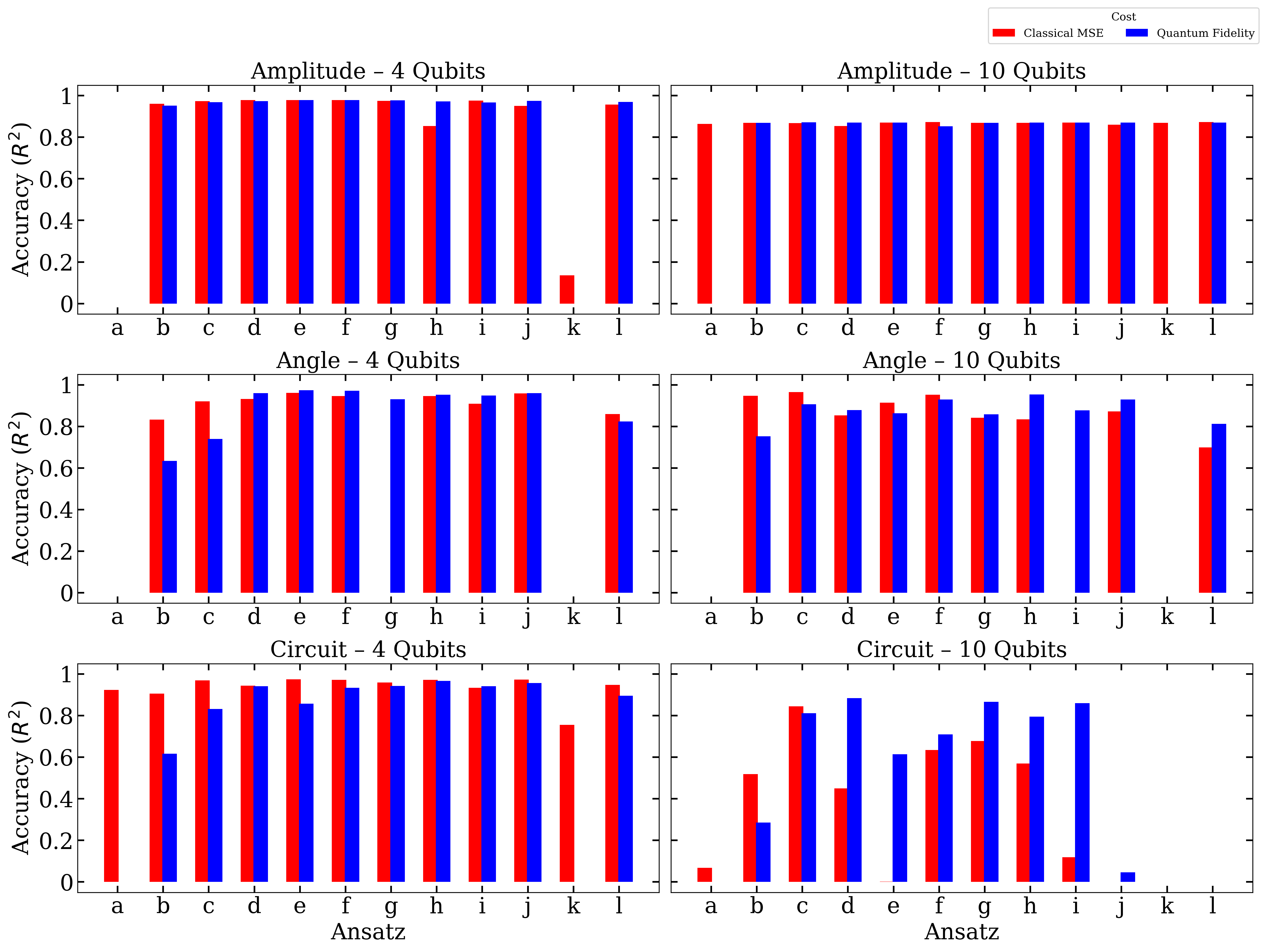}
    \caption{Mean $R^2$ score across the 5 runs for the various ansatz using different cost functions. Results are split across the encoding methods using both 4 and 10 qubits, and all models used 8 layers.}
    \label{fig:acc-cost}
\end{figure}

The cost function affects accuracy differently depending on the encoding method and number of qubits used, as shown in Figure \ref{fig:acc-cost} for 4 and 10 qubits at 8 layers. Amplitude encoding shows no clear trend between cost functions at either qubit count, whereas angle encoding favours the quantum cost function at both qubit counts. For circuit encoding, the classical cost function outperforms the quantum cost function at 4 qubits, but this reverses at 10 qubits. Here the quantum cost function performs better, although neither cost function achieves strong overall accuracy at this qubit count.

On average, a quantum cost function produces more accurate models, with a mean MSE of $0.02956 \pm 0.00212$ and mean $R^2$ of $-6.810 \pm 0.559$, compared to $0.05441 \pm 0.00649$ and $-13.216 \pm 1.693$ for the classical cost function. However, the best individual models tend to use a classical cost function: six of the top nine models (those achieving $R^2 > 0.983$) employed a classical cost function, compared to three using a quantum cost function. One possible explanation is that the classical cost function is MSE-based, which aligns directly with the MSE evaluation metric, potentially giving it an inherent advantage in optimising for the assessed performance measure.

\subsubsection*{Computational Cost}
\begin{figure}[htbp]
    \centering
    \includegraphics[width=1\textwidth]{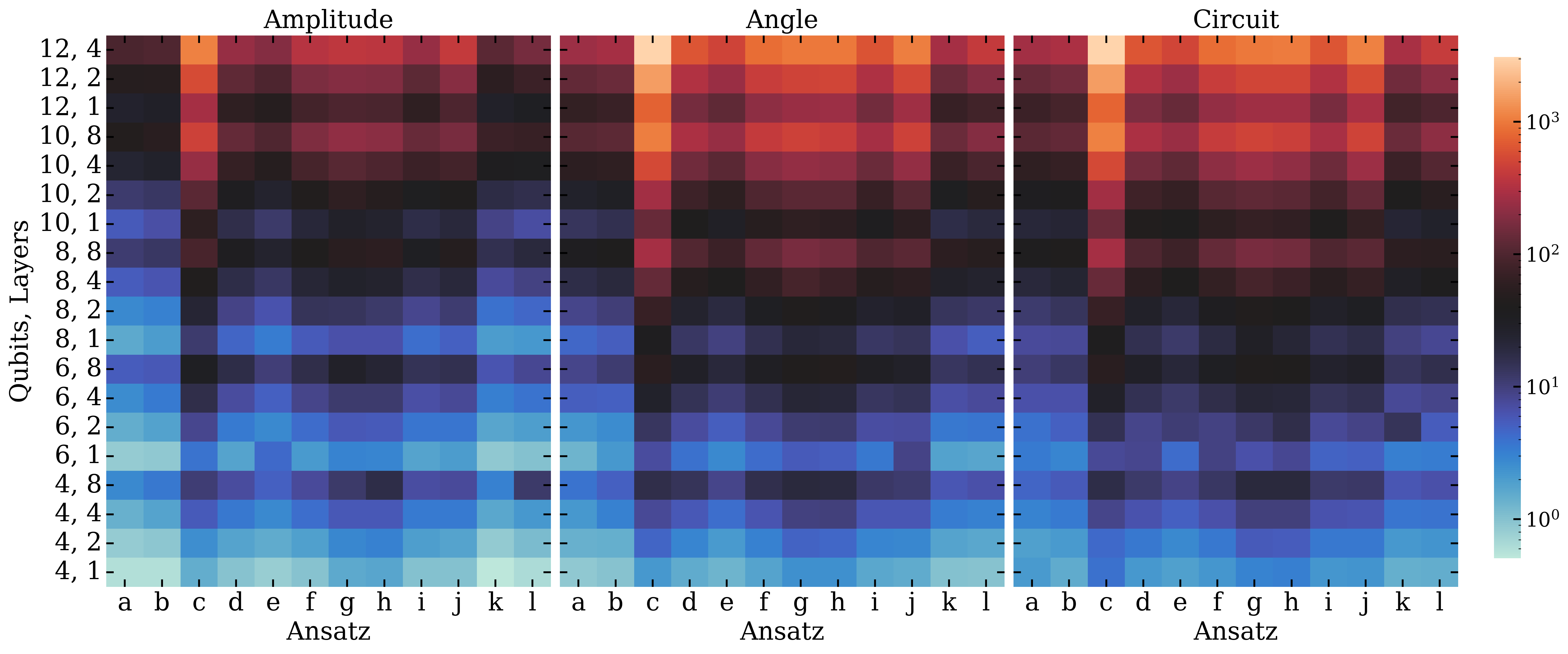}
    \caption{A heatmap showing the mean training time (s) across the 5 runs for the various ansatz, qubits and layers for models using a classical MSE cost function, split across the 3 encoding methods. The colour scale used is a logarithmic scale.}
    \label{fig:heat-run}
\end{figure}

The training time for individual models is considered to assess computational cost for training via classical simulation. All figures use the mean training time across the five runs. Figure \ref{fig:heat-run} presents a heatmap of training times across all evaluated ansatz variants, qubit counts, and layer depths for models  using the classical MSE cost function. A clear increase in training time is observed as both qubit count and circuit depth increase, although the magnitude of this increase varies between ansatz and encoding methods.

\begin{figure}[htbp]
    \centering
    \includegraphics[width=\textwidth]{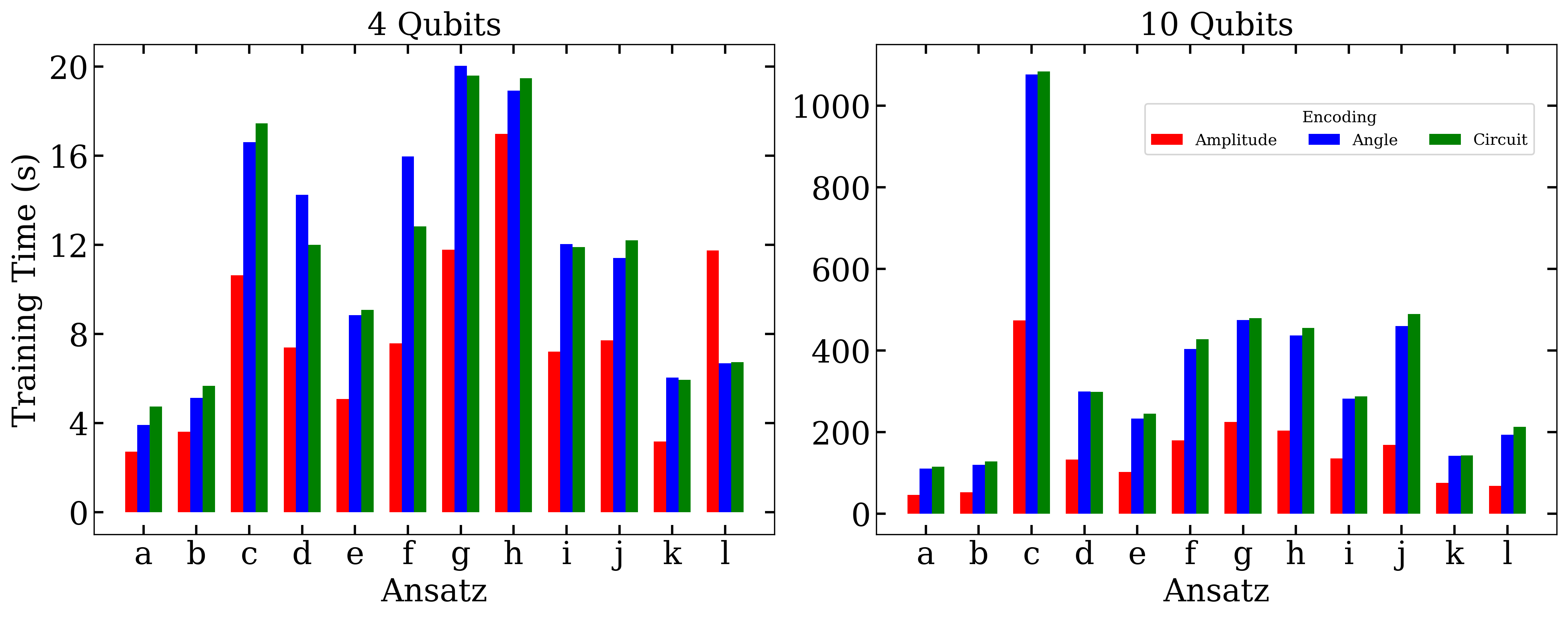}
    \caption{Mean training time (s) across the 5 runs for the various ansatz using different encoding methods. Results are split across models using both 4 and 10 qubits, and all models used 8 layers and a classical MSE cost function.}
    \label{fig:run-enc}
\end{figure}

Figure \ref{fig:run-enc} further illustrates the differences between encoding methods for selected four- and ten-qubit configurations. At 4 qubits, amplitude encoding is faster than both angle and circuit encoding, with angle and circuit encoding much closer to each other. At 10 qubits the separation is more pronounced: amplitude encoding is the fastest by a significant margin, and circuit encoding is consistently slower than angle encoding across all ansatz. Mean training times across all models are $50.48 \pm 5.09$ seconds for amplitude encoding, $126.27 \pm 13.45$ for angle encoding, and $132.21 \pm 13.68$ for circuit encoding. Ansatz (c) is noticeably slower than other ansatz at higher qubit counts.

\begin{figure}[htbp]
    \centering
    \includegraphics[width=1\textwidth]{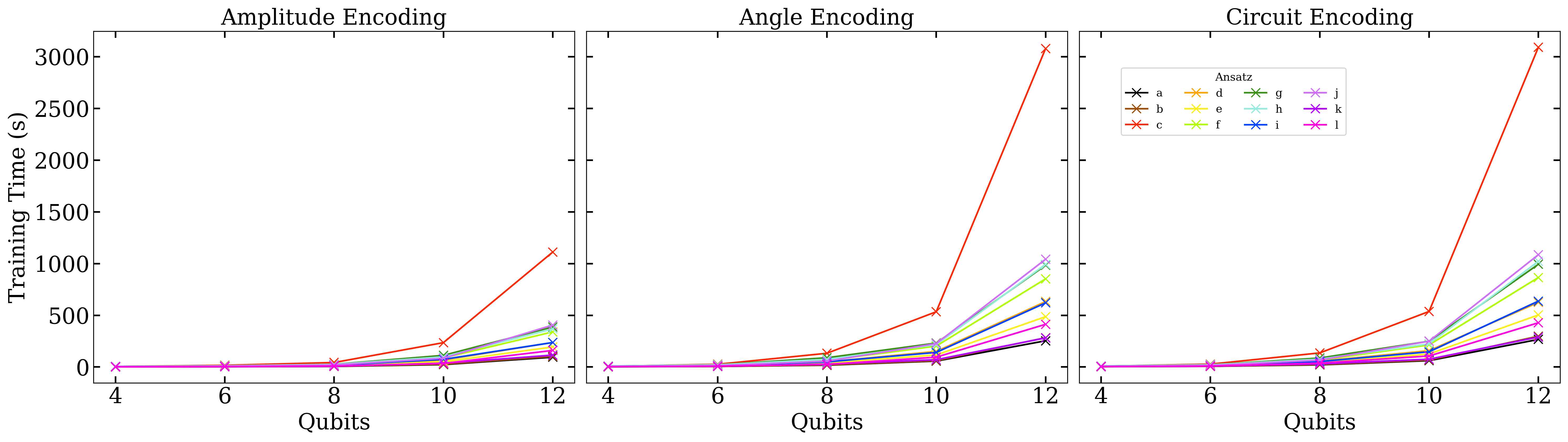}
    \caption{Mean training time (s) across the 5 runs for different number of qubits using the various ansatz. Results are split across models using the 3 encoding methods, and all models used 4 layers and a classical MSE cost function.}
    \label{fig:run-qubits}
\end{figure}

\begin{figure}[htbp]
    \centering
    \includegraphics[width=1\textwidth]{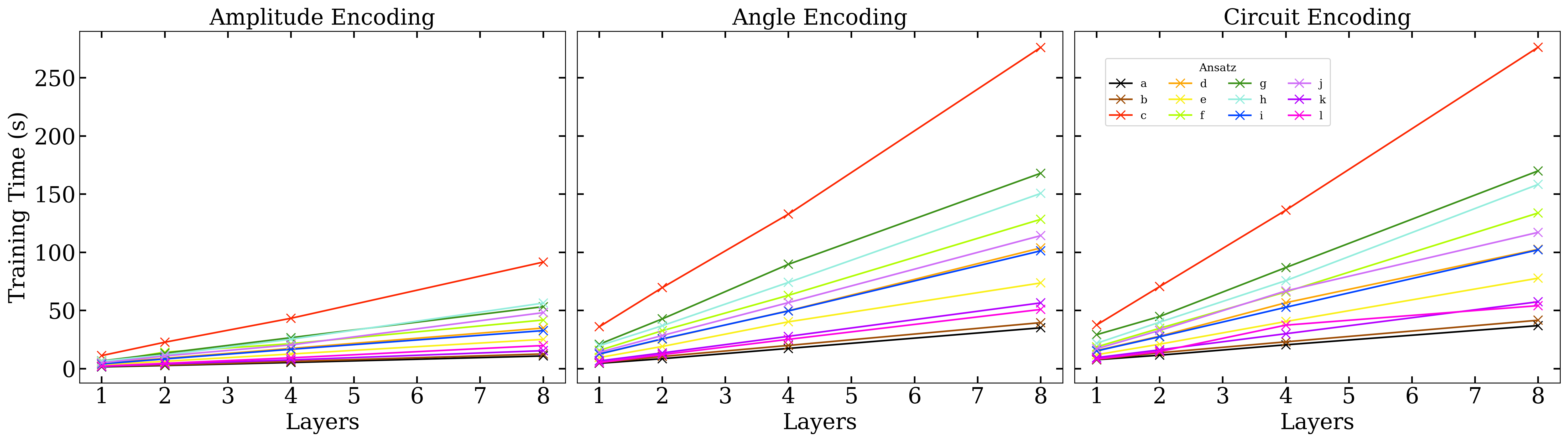}
    \caption{Mean training time (s) across the 5 runs for different number of layers using the various ansatz. Results are split across models using the 3 encoding methods, and all models used 8 qubits and a classical MSE cost function.}
    \label{fig:run-layers}
\end{figure}

The scaling behaviour with respect to qubit count and layer depth is then examined separately in Figures \ref{fig:run-qubits} and \ref{fig:run-layers}. As shown in Figure \ref{fig:run-qubits}, training time increases exponentially with qubit count across all encoding methods, while Figure \ref{fig:run-layers} shows a linear increase with layer count, with the gradient varying between ansatz.

\begin{figure}[htbp]
    \centering
    \includegraphics[width=1\textwidth]{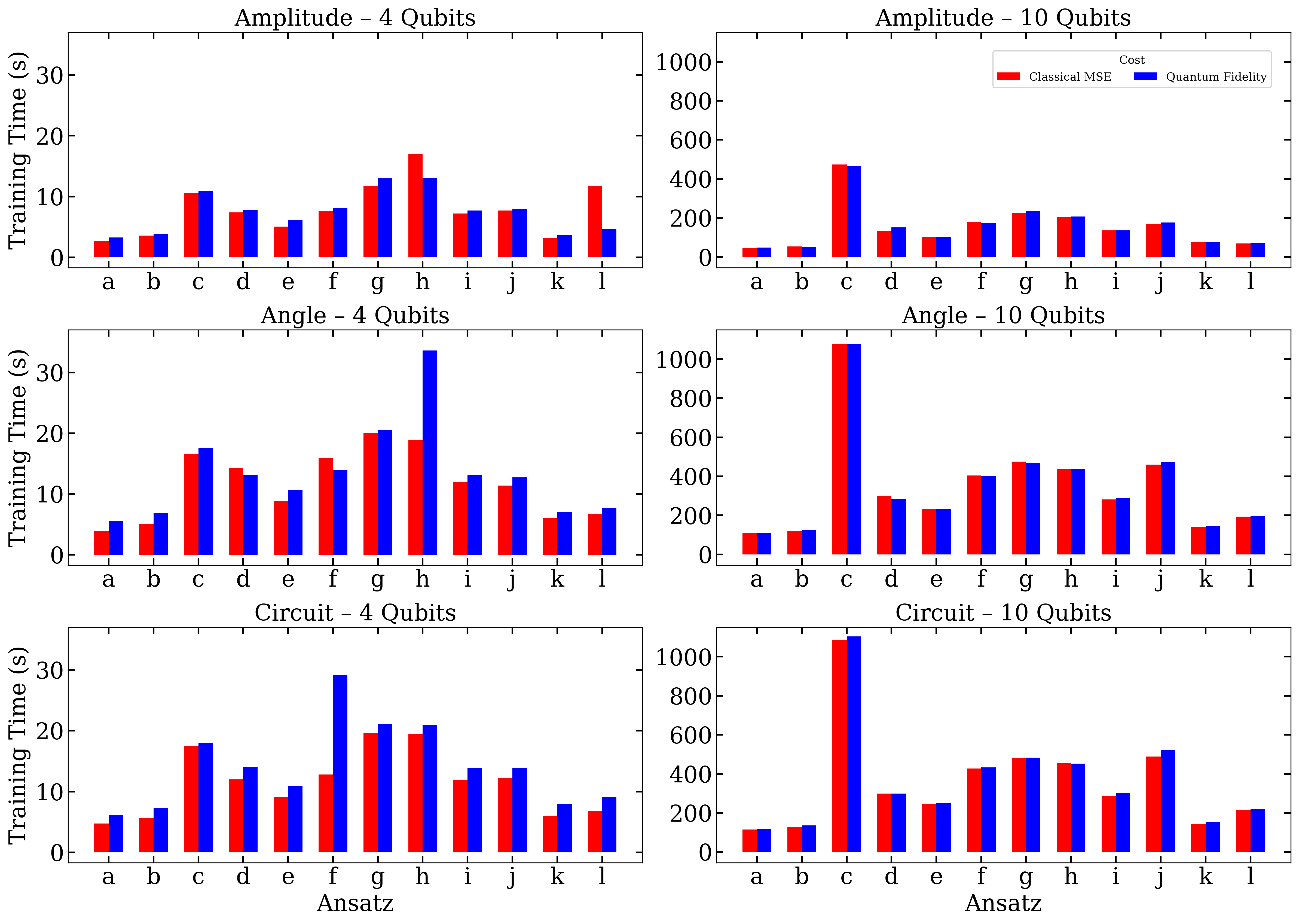}
    \caption{Mean training time (s) across the 5 runs for the various ansatz using different cost functions. Results are split across the encoding methods using both 4 and 10 qubits, and all models used 8 layers.}
    \label{fig:run-cost}
\end{figure}

The cost function also affects training time, as shown in Figure \ref{fig:run-cost}. At both 4 and 10 qubits, the quantum cost function takes longer to train than the classical cost function across almost all encoding methods and ansatz, though the proportional difference is smaller at 10 qubits than at 4 qubits.

\subsubsection*{Time to convergence}
\begin{figure}[htbp]
    \centering
    \includegraphics[width=1\textwidth]{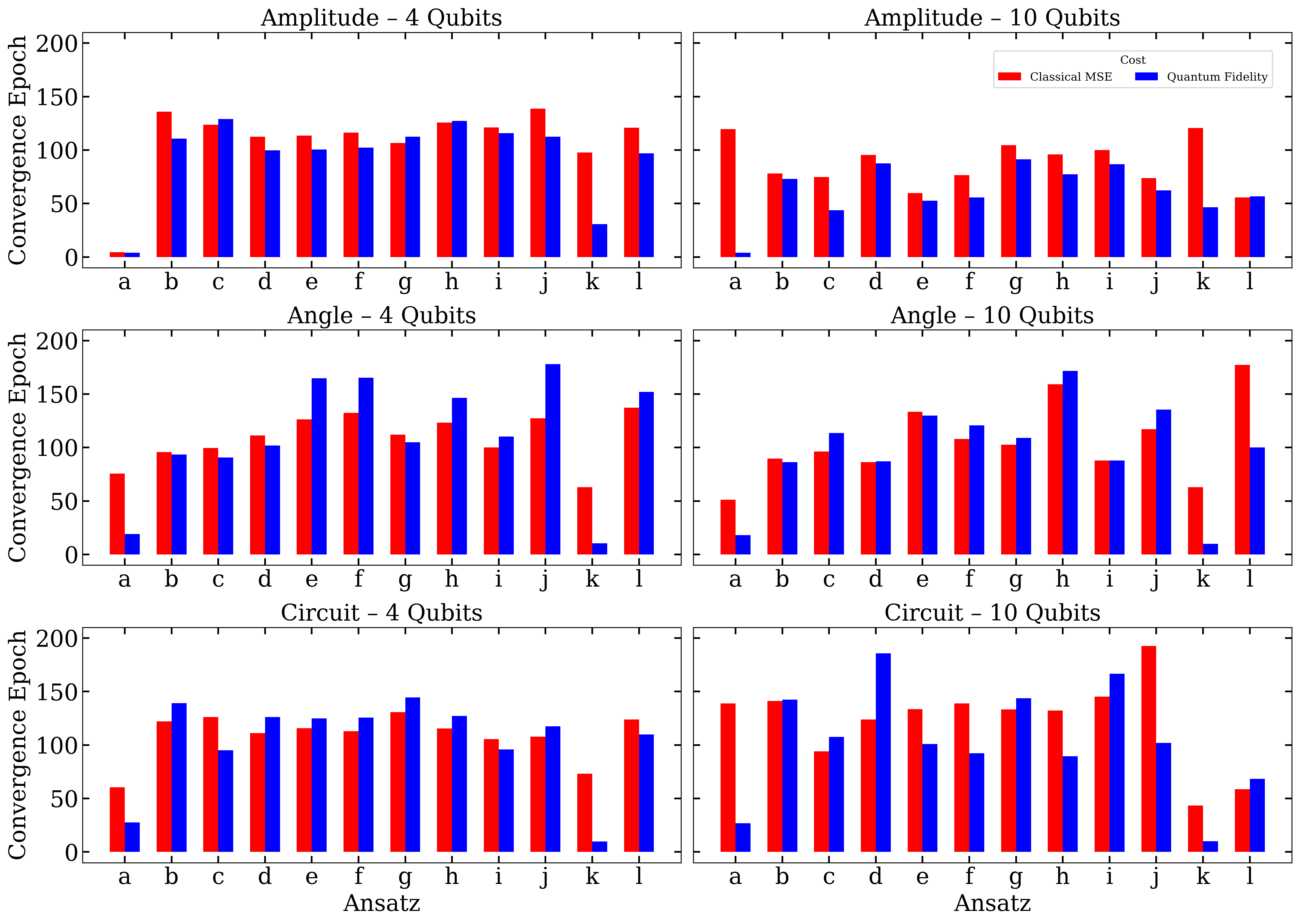}
    \caption{Mean convergence epoch across the 5 runs for the various ansatz using different cost functions. Results are split across the encoding methods using both 4 and 10 qubits, and all models used 8 layers.}
    \label{fig:conv-cost}
\end{figure}

To assess the time to convergence of the models, the convergence epoch was calculated for each configuration and averaged across the corresponding models. The mean convergence epoch seems to decrease with increasing qubit count, from $94.83 \pm 2.49$ at 4 qubits to $76.85 \pm 2.87$ at 12 qubits. However, this apparent trend does not account for poorly performing models that do not converge. Configurations with $R^2<0$ have an average convergence epoch of $54.24 \pm 1.72$, while successful models with $R^2 \in [0.8,1]$ have an average of $85.22 \pm 1.10$. Poorly performing models therefore tend to plateau at earlier epochs, reducing the overall mean and creating the appearance of a relationship between increasing qubit count trend and fast convergence. Similar distortion is observed for ansatz and layer count. Trends relating to the cost function, shown in Figure \ref{fig:conv-cost}, appear independent of this effect, as the data shown uses 8 layers where models primarily converged. For amplitude encoding, a quantum fidelity cost function converges faster across almost all ansatz at both qubit counts. For angle encoding, the classical MSE cost function converges faster at 4 qubits, with the gap narrowing at 10 qubits. For circuit encoding, the classical cost function converges faster across most ansatz at both qubit counts, though when the quantum cost function does converge faster, it does so by a larger margin.

\subsection{Impact of Quantum Hardware Noise on Accuracy}
To assess how hardware noise affects prediction accuracy, models trained on the noiseless classical simulator are evaluated for inference across three execution environments: the noiseless classical simulator, the Amazon Braket SV1 simulator with readout noise applied, and the IQM Emerald quantum device. This comparison enables decomposition of the total noise contribution into readout noise and broader hardware noise effects introduced by real quantum device.

\begin{figure}[htbp]
    \centering
    \includegraphics[width=1\textwidth]{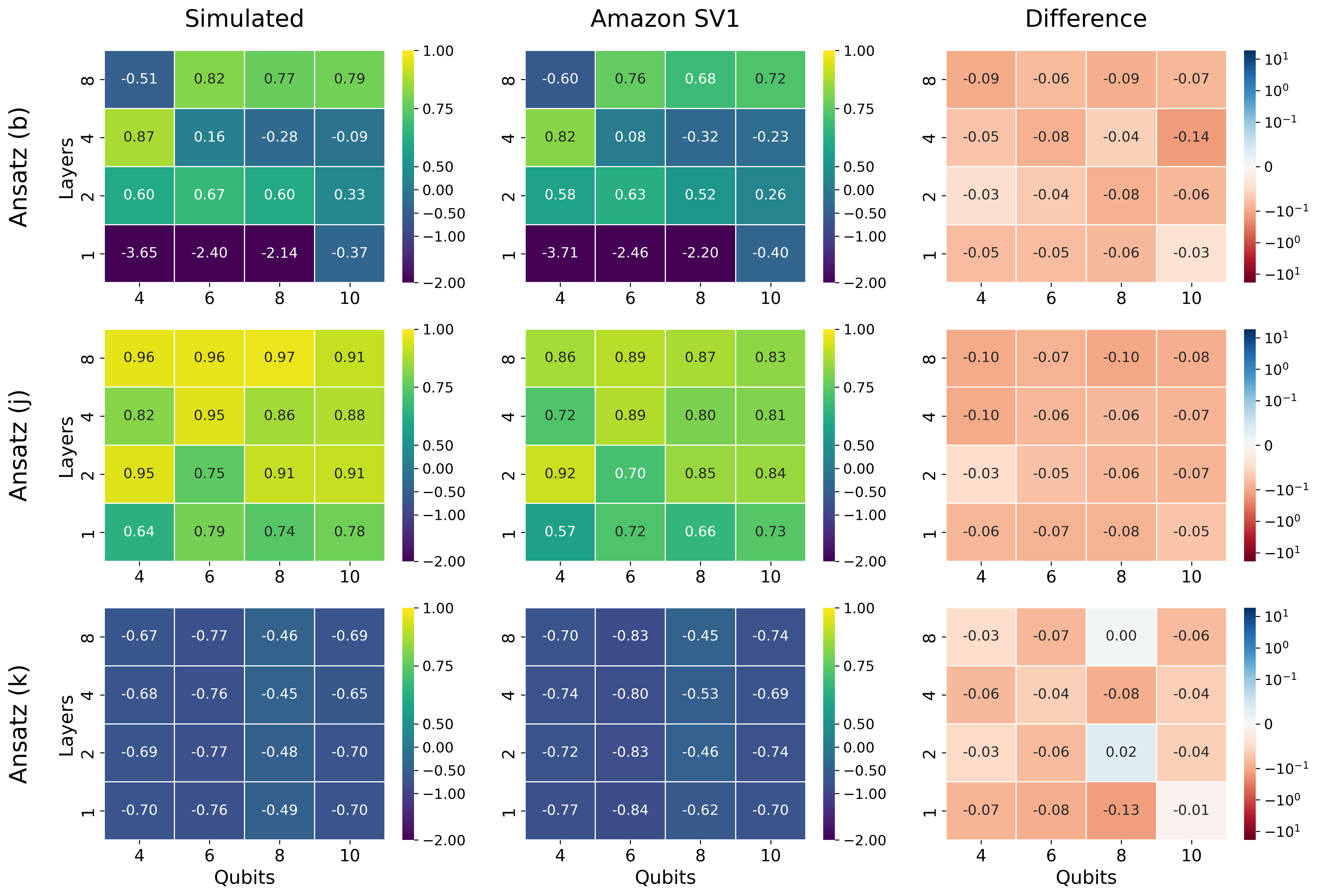}
    \caption{Heatmaps displaying the $R^2$ scores across the three hardware ansatz variants, qubit counts, and layer configurations for models using angle encoding and a classical MSE cost function. Results show the configurations implemented on the noiseless classical simulator (Simulated), the Amazon Braket SV1 simulator (Amazon SV1), alongside the corresponding change in $R^2$ score between the two environments (Difference).}
    \label{fig:sv1-hard}
\end{figure}

Figure \ref{fig:sv1-hard} shows the $R^2$ scores across the ansatz, qubit counts, and layer configurations for the comparison between the classical and SV1 simulator. The introduction of readout noise produces a minor and broadly uniform degradation in prediction accuracy across all configurations, with $R^2$ scores decreasing by an average of approximately 0.06 compared to the noiseless baseline. 

\begin{figure}[htbp]
    \centering
    \includegraphics[width=1\textwidth]{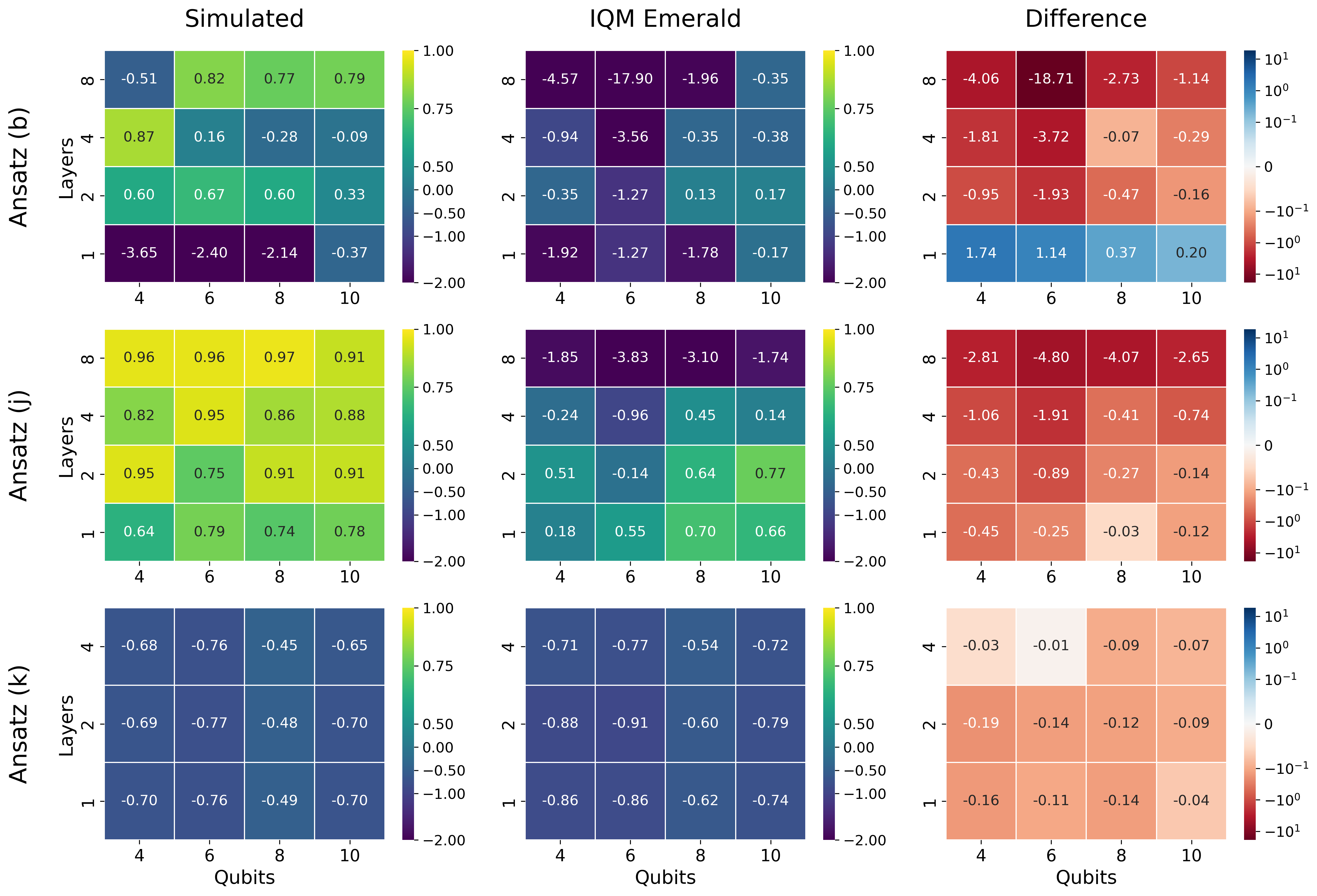}
    \caption{Heatmaps displaying the $R^2$ scores across the three hardware ansatz variants, qubit counts, and layer configurations for models using angle encoding and a classical MSE cost function. Results show the configurations implemented on the noiseless classical simulator (Simulated), the IQM Emerald quantum device (IQM Emerald), alongside the corresponding change in $R^2$ score between the two environments (Difference).}
    \label{fig:emerald-hard}
\end{figure}

Figure \ref{fig:emerald-hard} shows the $R^2$ scores for the comparison between the classical simulator and IQM Emerald device. Execution on the IQM Emerald device produces a considerably more substantial decline in accuracy, with $R^2$ scores decreasing by an average of approximately 1.25, reflecting the additional contribution of gate errors and decoherence beyond readout noise alone. 

Several patterns are observable across the IQM Emerald results. Ansatz (b) at a single layer shows a marginal improvement over the noiseless simulated baseline across all qubit counts, representing the only configuration where hardware execution consistently produces higher $R^2$ values than the baseline. However, the resulting $R^2$ scores remain negative, indicating that the apparent improvement does not translate into strong predictive performance.

Ansatz (j) at one and two layers produces the strongest results on the IQM Emerald device, achieving a best $R^2$ of 0.77, demonstrating that meaningful forecasting performance can be retained under real hardware noise when relatively shallow circuit configurations are used. Accuracy degrades further as the number of layers increases, a trend consistent across ansatz (b) and (j), suggesting that deeper circuits are more susceptible to accumulated hardware noise and device-level errors. 

For ansatz (k), the additional degradation of accuracy beyond the SV1 results is comparatively marginal. However, this result must be contextualised by ansatz (k)'s inherently weak baseline performance. Whilst no clear trend is observed with respect to qubit count, six-qubit configurations across ansatz (b) and (j) show notably poorer performance than the corresponding configurations at other qubit counts, particularly at greater circuit depth. 

Overall, the results across the three execution environments reveal a clear hierarchy of noise impact. Readout noise introduced through SV1 produces modest and relatively uniform degradation, while execution on the IQM Emerald device results in substantially greater reductions in $R^2$ for many configurations. Nevertheless, selected shallow configurations, particularly for ansatz (j), retain  meaningful predictive capability under real quantum hardware conditions.

\section{Discussion} \label{sec:Discuss}
\subsection*{Ansatz Design and Circuit Composition}
The \textit{mean} and \textit{best individual performance} metrics serve complementary purposes in evaluating model performance. \textit{Mean} performance reflects the consistency with which a configuration converges to a valid solution, which is the relevant metric when reliability is prioritised over marginal accuracy gains. \textit{Best individual performance} reflects the achievable ceiling of a configuration, which is relevant when peak accuracy justifies the additional computational resources required to attain it. The two metrics do not always coincide for a specific configuration; ansatz (f) ranks eighth for mean accuracy but second for best individual accuracy, illustrating that a configuration can have a high ceiling alongside inconsistent convergence. Notably, the top five configurations by best individual $R^2$ — ansatz (d), (e), (f), (h), and (j) — achieve very similar scores, suggesting that the achievable accuracy ceiling is relatively uniform across a range of well-chosen ansatz. The significance of this accuracy ceiling is further contextualised by comparison with a classical BiLSTM baseline. Ten of the twelve ansatz configurations outperformed the constrained BiLSTM baseline ($R^2 = 0.9529$) on their best individual model, which was trained under equivalent data conditions, indicating that well-configured QNNs can surpass classical models operating under the same constraints. Furthermore, the eight best-performing ansatz configurations outperform the unconstrained BiLSTM ($R^2 = 0.9805$), despite the latter benefiting from a substantially longer lookback window and fully overlapping windows. This indicates that the strongest QNN configurations are not only competitive with, but also capable of exceeding classical models that operate without such restrictions. These findings suggest that the data constraints imposed by the quantum architecture do not necessarily preclude strong predictive performance. Whilst a controlled quantum advantage study is beyond the scope of this work, these results provide encouraging preliminary evidence of the potential for QNNs in financial time series forecasting.

Among these well-chosen ansatz, the composition affects performance more than its raw scale. Ansatz (c) has the most parameters (for models with six or more qubits), the most two-qubit gates, and the greatest circuit depth of any ansatz tested, yet ranks only fifth on mean accuracy and ninth on best individual accuracy. Conversely, the two best individual performers, ansatz (j) and (f), contain no parameterised two-qubit gates suggesting that such gates are not a prerequisite for peak accuracy, although both circuits still contain non-parameterised entangling gates. The success of an ansatz therefore depends on the choice and arrangement of gates not just their quantity, as this determines the circuit's ability to explore the relevant region of the Hilbert space and represent the target function.

A consistent indicator across the data is the presence of single-qubit parameterised rotation gates, as all top four ansatz by mean accuracy include these. This is highlighted most clearly by comparing ansatz (a) and (i), which share the same two-qubit gate structure but differ in that (i) adds single-qubit rotations. Ansatz (i) performs significantly better on both metrics, indicating that the rotation gates themselves contribute to performance rather than merely correlating with it. The best performing ansatz generally consist of a balance of both single- and two-qubit gates. As single-qubit parameterised rotations provide independent control over individual qubit states in arbitrary directions on the Bloch sphere, the circuit has greater flexibility to shape its output distribution than if only two-qubit gates are used.

Whilst not the dominant contributor to accuracy, parameter count may still be relevant. Ansatz (h), which achieves the highest mean accuracy and the only positive mean $R^2$, contains the greatest number of parameters and two-qubit gates among the top four mean performers. More broadly, while raw accuracy does not scale monotonically with parameter count, higher parameter models are more likely to converge to a valid solution. This also helps explain why increasing the number of layers improves accuracy more consistently in comparison to increasing the number of qubits, where although the parameter count still increases, additional structural changes to the circuit are also introduced. Adding layers introduces parameters without altering other structural properties of the circuit, isolating the benefit of increased expressiveness from the complications that additional qubits introduce.

Ansatz (k)'s position as the worst performer by a considerable margin on both metrics reflects its lack of two-qubit gates and therefore any entangling capability. A source of entanglement is evidently required for a circuit to converge to a valid solution. However, this entanglement should be considered a system-level property rather than an ansatz-level one. Ansatz (a) and (k), the two ansatz with minimal entangling capability, frequently fail to converge, although, under a classical MSE cost function, they can converge when paired with amplitude or circuit encoding. Both encoding methods introduce entanglement through their encoding circuits, evidently compensating for the ansatz's limitations. Angle encoding introduces no such entanglement, and the same minimally entangled ansatz consequently fail to converge under it. Total circuit entanglement therefore matters more than where in the circuit it originates, nevertheless convergence alone is not sufficient, as ansatz (a) and (k) still perform poorly even when they do converge. The entanglement supplied by the encoding method can support the model, but is not enough to compensate for a complete lack of entanglement in the ansatz. It should also be noted that this pattern does not arise under a quantum cost function.

\subsection*{Optimisation Landscape and Cost Function Trade-offs}
Barren plateaus may partly account for several of the qubit scaling trends observed in this study, as their effect intensifies exponentially with qubit count. Although measures were taken to mitigate this during training, the effect cannot be ruled out as a contributor to the broad accuracy declines seen as qubits increase, particularly for circuit encoding where the decline is most pronounced. Why this effect is so much stronger for circuit encoding than for the other methods is unclear and warrants further investigation. Barren plateau susceptibility may also help explain the cost function trade-off, where the quantum fidelity cost function produces better mean accuracy than the classical MSE cost function, but the classical cost function produces the best individual models. One possible interpretation is that the classical cost function has a higher ceiling but greater variance, encountering barren plateaus more frequently and failing more often while occasionally finding higher-quality optima. The quantum cost function may correspondingly produce a tighter distribution of outcomes with fewer failures but a lower peak. This remains speculative and would benefit from a dedicated investigation comparing loss landscape geometry under each cost function.

\subsection*{Computational Considerations}
Beyond accuracy, training time scales predictably with circuit structure, and the observed trends match theoretical expectations. Training time scales linearly with circuit depth and, by extension, parameter count, evidenced by the linear increase with layer count in Figure \ref{fig:run-layers}. Adding layers introduces parameters and circuit depth without altering other circuit properties, isolating the circuit depth effect. Gradients for individual ansatz in the layer plot follow from each ansatz's parameter and depth scaling (Table \ref{tab:ansatz}), and ansatz (c)'s disproportionate slowness at high qubit counts reflects its quadratic parameter growth. Training time scales exponentially with qubit count across all encoding methods, reflecting the exponential growth of the underlying Hilbert space rather than parameter count alone. Differences between encoding methods follow the same logic; amplitude encoding's exponential data capacity gives it a substantial training time advantage at higher qubit counts, while circuit encoding's larger encoding circuit makes it slightly slower than angle encoding despite both methods evaluating the same amount of data. The cost function introduces an additional, structural training time difference with the quantum fidelity cost function consistently slower than classical MSE, but the absolute gap does not grow with qubit or parameter count. This indicates that the difference arises from how each cost function evaluates the circuit, notably the quantum measurement step required for fidelity, rather than from circuit complexity.

Convergence epoch trends require more careful interpretation. The apparent decrease in convergence epoch with qubit count is an artifact of non-convergence rather than a genuine acceleration of training. Models that fail to converge stop improving early, registering low convergence epochs that pull the mean downward. Configurations associated with poorer accuracy, such as high qubit counts, weaker ansatz, and lower layer counts, therefore appear to converge faster simply because their failed runs dominate the average. The cost function trends shown in Figure \ref{fig:conv-cost} are an exception, as the data there is restricted to 8-layer models which largely converged, suggesting those trends reflect genuine differences in optimisation dynamics between the classical MSE and quantum fidelity cost functions.

\subsection*{Hardware Noise and Near-Term Deployment}
The SV1 results demonstrate that readout errors account for only a minor fraction of the total performance loss, with $R^2$ scores declining by an average of approximately 0.06 across all configurations. Notably, this degradation is broadly uniform across ansatz variants, and qubit and layer counts, which is consistent with the experimental design. Since circuit output is averaged across four qubits regardless of the total qubit count, the readout noise contribution remains approximately constant across configurations.

Execution on the IQM Emerald device reveals a considerably larger noise contribution from gate errors and decoherence. Several trends emerge across the IQM Emerald results. Ansatz (k) exhibits the smallest additional degradation beyond the SV1 baseline, attributable to its absence of two-qubit gates. Single-qubit gates operate at substantially higher fidelity on current hardware, and since ansatz (k) consists exclusively of these gates, gate errors accumulate at a much lower rate. However, this represents a direct trade-off, as the same lack of two-qubit gates that provides hardware resilience also eliminates entangling capability, which as previously established earlier in the discussion, results in poor predictive performance. Ansatz (b) shows greater degradation than ansatz (j) despite seemingly having a lower circuit depth, a result of its use of controlled-X rotation (CRX) gates which are non-native to the IQM Emerald device. In actuality, a gate decomposition replaces each CRX gate with four single-qubit gates and two CNOT gates, which substantially increases the circuit depth, reducing the effective gate fidelity and accelerating error accumulation. This is exemplified further by the modified variant of ansatz (d), where the extent of non-native gate decomposition resulted in circuit depths that exceeded the capacity of the IQM Emerald circuit compiler entirely beyond four layers. This highlights a broader practical constraint for near-term hardware deployment, by which ansatz selection must account not only for circuit depth and expressivity in simulation but also for native gate compatibility with the target device, as gate decomposition can render otherwise viable circuits impractical. Furthermore, the apparent improvement observed for single-layer ansatz (b) across qubit counts is misleading. Inspection of individual forecasts reveals that the noiseless simulated predictions are already poor for these configurations, and the large amount of hardware noise experienced for ansatz (b) incidentally shifts predictions closer to the true data. This represents a coincidental alignment of a poor forecast and high noise rather than a genuine performance gain. Accuracy generally degrades as layer count increases, which is consistent with the polynomial growth of two-qubit gate count with layers. The gate errors compound with circuit depth, as the lower fidelity of two-qubit gates means each additional layer introduces a compounding source of error. This demonstrates how the increased expressivity gained by increasing the number of layers must be balanced appropriately with the level of degradation associated with it, when implemented on quantum hardware. This trend is not observed in ansatz (k), where the absence of two-qubit gates limits error compounding. No equivalent trend is observed with qubit count, likely because the growth in circuit depth with additional qubits is less significant than the growth with additional layers. The disproportionately poor performance of six-qubit configurations across ansatz (b) and (j) may reflect interference between parallel circuits. Various parallel circuit configurations are used for the different qubit counts in each quantum execution to reduce execution costs, and the configuration for 6 qubits may introduce greater inter-circuit noise than other configurations, though this remains speculative and warrants further investigation.

Taken together, these results indicate that hardware noise currently represents a significant barrier to practical standalone QNN deployment for financial time series forecasting, with only a small subset of low layer configurations retaining meaningful predictive capability on the IQM Emerald device. The choice of gates, layer count, and ansatz structure must be carefully considered when targeting real hardware, as the optimal configuration for noiseless simulation does not necessarily translate to the best hardware performance. It should also be noted that the noise sensitivity observed here applies to standalone QNN execution. In hybrid quantum-classical architectures, where the QNN layer is embedded within a larger classical model, the impact of hardware noise may be partially absorbed or compensated by the surrounding classical components \cite{422Ahmed}, potentially offering a more noise-resilient deployment pathway for near-term devices. Nevertheless, as gate fidelities and qubit coherence times continue to improve with advances in quantum hardware, a broader range of configurations are expected to produce useful results, particularly those employing deeper circuits and more expressive ansatz. The findings presented here provide a baseline characterisation of noise sensitivity across circuit configurations, offering practical guidance for hardware-aware QNN design on NISQ devices.

\section{Conclusions and Future Work} \label{sec:Conc}
By systematically interchanging QNN components across 1368 distinct configurations, this study provides a comprehensive evaluation of how architectural choices affect QNN performance for financial time series forecasting. The best individually performing model when forecasting the GBP/USD spot exchange rate achieved an $R^2$ score of 0.98495 using angle encoding with ansatz (j), 4 qubits, 8 layers, and a classical MSE cost function, while amplitude encoding achieved the greatest mean accuracy across all configurations. Comparison with a classical BiLSTM baseline established under equivalent data conditions shows that ten of the twelve ansatz configurations outperform the constrained baseline, with the eight best-performing configurations also exceeding an unconstrained BiLSTM with access to a longer lookback window, providing encouraging evidence of the potential for QNNs in financial forecasting. Beyond these headline results, the analysis yielded several insights into QNN behaviour. Gate selection and arrangement proved more critical to model success than raw parameter count, though once an effective ansatz structure is established, increasing parameter count through additional layers tends to improve performance consistently. Contrary to typical expectations, smaller qubit counts generally produced better results, a finding attributed in part to the exponential data capacity of amplitude encoding and the potential influence of barren plateaus at higher qubit counts. The study also establishes that entanglement is a system-level property of the full circuit rather than solely an ansatz-level one, with encoding methods capable of partially compensating for limited ansatz entanglement under certain conditions. The hardware noise investigation conducted on the IQM Emerald quantum device revealed that gate errors and decoherence currently represent a significant barrier to practical deployment, producing substantially greater accuracy degradation than readout noise alone. These findings provide practical guidance for hardware-aware QNN design and establish a baseline characterisation of noise sensitivity across circuit configurations on near-term devices.

This work also highlights several directions for future research.  Whilst QNNs have shown promise in limited data scenarios, processing large-scale datasets remains a significant challenge under current NISQ constraints, though continued hardware advances may address this limitation. Barren plateaus present another persistent challenge for PQCs, though strategies to mitigate this effect are an active area of research \cite{352Puig}\cite{353Zhang}\cite{354Liu}\cite{355Zambrano}\cite{356Sack}. As QNNs form the foundations for more advanced quantum models, such as QCNNs, QRNNs, and QLSTMs, the configurations identified here as optimal could be evaluated within these more complex models to assess whether the findings generalise. As gate fidelities and coherence times improve with continued advances in quantum hardware, the hardware barriers identified in this study are expected to diminish, enabling a broader range of configurations to produce practically useful results. Although substantial further research is required  before QNNs can achieve a level of practical utility comparable to classical models, the outlook remains promising, particularly in light of the rapid advances in quantum hardware.

\backmatter

\bmhead{Acknowledgements}
This project was partially funded and supported by the UK National Quantum Computer Centre [NQCC200921], which is a UKRI Centre and part of the UK National Quantum Technologies Programme (NQTP). J.W. acknowledges support from the Kingston University PhD Studentship.

\section*{Declarations}
\begin{itemize}
\item Funding
This project was partially funded and supported by the UK National Quantum Computer Centre [NQCC200921], which is a UKRI Centre and part of the UK National Quantum Technologies Programme (NQTP).

\item Competing interests 
The authors declare no competing interests.

\item Ethics approval and consent to participate
Not applicable

\item Consent for publication
Not applicable

\item Data availability 
The datasets used in this study are available at https://www.bankofengland.co.uk/boeapps/database/.

\item Materials availability
Not applicable

\item Code availability 
Code is available from the corresponding author upon reasonable request and will be made publicly available on GitHub upon publication.

\item Author contribution
J.W. conceived and designed the study. J.W. developed the model, implemented the code, conducted the experiments, analysed the results, generated the figures, and wrote the original draft. X.L., D.M., and R.N. supervised the project. X.L. and D.M. secured funding. X.L., D.M., and R.N. reviewed and edited the manuscript.

\end{itemize}

\bibliographystyle{unsrtnat}
\bibliography{bibliography}

\end{document}